\documentclass[manuscript,screen]{acmart}

\AtBeginDocument{%
  }

\setcopyright{acmlicensed}
\copyrightyear{2025}
\acmYear{2025}
\acmDOI{XXXXXXX.XXXXXXX}

\newtheorem{definition}{Definition}

\usepackage{algorithm}
\usepackage{algorithmic}
\usepackage{makecell}
\DeclareMathOperator*{\argmax}{argmax}
\usepackage{mathtools}
\usepackage{subcaption}
\usepackage{placeins}
\usepackage{tabularx}
\usepackage{wrapfig}
\usepackage{hyperref}
\usepackage{hyperxmp}
\usepackage{orcidlink}
\usepackage[most]{tcolorbox} 

\newcolumntype{L}{>{\raggedright\arraybackslash}X}
\newcolumntype{C}{>{\centering\arraybackslash}X}

\usepackage[T1]{fontenc}
\usepackage[font=small,labelfont=bf,tableposition=top]{caption}
\DeclareCaptionLabelFormat{andtable}{#1~#2  \&  \tablename~\thetable}

\begin{document}

\title{Network-Constrained Policy Optimization for Adaptive Multi-agent Vehicle Routing}

\author{Fazel Arasteh\,\orcidlink{0000-0002-4551-0771}}
\authornote{These authors contributed equally to this research.\\
The code for this research is publicly available at \url{https://github.com/Arianhgh/HHAN}}
\email{fazelara@eecs.yorku.ca}
\affiliation{%
  \institution{York University}
  \country{Canada}
}

\author{Arian Haghparast\,\orcidlink{0009-0001-2277-8038}}
\authornotemark[1]
\email{arianhgh@my.yorku.ca}
\affiliation{%
  \institution{York University}
  \country{Canada}
}

\author{Manos Papagelis\,\orcidlink{0000-0003-0138-2541}}
\email{papaggel@eecs.yorku.ca}
\affiliation{%
  \institution{York University}
  \country{Canada}
}

\renewcommand{\shortauthors}{Arasteh, Haghparast, Papagelis}

\begin{abstract}

Traffic congestion in urban road networks is marked by longer trip times and higher emissions, especially during peak periods. While the \textsc{Shortest Path First} (SPF) algorithm is optimal for a single vehicle in a static network, it performs poorly in dynamic, multi-vehicle settings, often worsening congestion by routing all vehicles along identical paths. We address dynamic vehicle routing through a \textbf{multi-agent reinforcement learning (MARL)} framework for coordinated, network-aware fleet navigation. We first propose \textsc{Adaptive Navigation} (AN), a decentralized MARL model where each intersection agent provides routing guidance based on (i) local traffic and (ii) neighborhood state modeled using \textbf{Graph Attention Networks} (GAT). To improve scalability in large networks, we further propose \textsc{Hierarchical Hub-based Adaptive Navigation} (HHAN), an extension of AN that assigns agents only to key intersections (\textit{hubs}). Vehicles are routed hub-to-hub under agent control, while SPF handles micro-routing within each hub region. For hub coordination, HHAN adopts \textbf{centralized training with decentralized execution (CTDE)} under the \textbf{Attentive Q-Mixing (A-QMIX)} framework, which aggregates asynchronous vehicle decisions via attention. Hub agents use flow-aware state features that combine local congestion and predictive dynamics for proactive routing. Experiments on synthetic grids and real urban maps (Toronto, Manhattan) show that AN reduces average travel time versus SPF and learning baselines, maintaining 100\% routing success. HHAN scales to networks with hundreds of intersections, achieving up to \textbf{15.9\% improvement} under heavy traffic. These findings underscore the power of \textbf{network-constrained MARL} for scalable, coordinated, congestion-aware routing in intelligent transportation systems.
\end{abstract}



\received{N/A}
\received[revised]{N/A}
\received[accepted]{N/A}

\maketitle
\section{Introduction}
\label{chap:introduction}

\subsection{Motivation}
Traffic congestion in urban road networks is a condition characterized by longer trip times, increased air pollution, and driver frustration. Different factors like rush hours, traffic incidents, road maintenance work and bad weather conditions can contribute to the traffic congestion.
While construction of new road infrastructure is an expensive solution, the emergence of new technologies like widely available internet connection and GPS data can allow for more economical algorithmic traffic flow optimizations \cite{AV_Review_DL_for_traffic_flow_prediction}. Currently, services like Google Maps\footnote{https://maps.google.com/} and Waze\footnote{https://www.waze.com/} help people with route planning mainly relying on a variant of the popular Shortest Path First (SPF) algorithm \cite{dijkstra1959note}. Mostly known as the Dijkstra algorithm, SPF is a routing algorithm in which a router computes the shortest path between a pair of nodes in a network.

\subsection{Current approaches and limitations}
In a static network and for a single vehicle, the SPF algorithm is optimal. However, road network conditions are not always static. In a dynamic road network, the SPF path between an origin and a destination is harder to compute due to variable traffic conditions. The main approach to address this issue is to recursively break down the problem and estimate the travel time for smaller road segments, where the traffic conditions remain unchanged. This is usually referred to as the {\em traffic prediction problem}. Several methods have been proposed to address the {\em traffic prediction problem} of a road segment, ranging from classical time series prediction methods, such as historical average and autoregressive integrated moving average (ARIMA) models, to machine learning methods, such as Support Vector Regression and Random Forest. More recently, deep learning methods have been proposed to address the traffic prediction problem \cite{yin2021deep,tedjopurnomo2020survey}. Still, the estimated travel times, specifically long-term predictions, may be inaccurate, rendering the global SPF algorithm to be sub-optimal at times.
Another drawback of the SPF algorithm is that in multiple-vehicles scenarios, it will route every single vehicle through the currently available shortest path. As a result, due to the limited capacity of roads, the current shortest path gets quickly congested. In other words, SPF is short-sighted and causes congestion by naively sending every vehicle through the same shortest path. Other methods, such as probabilistic dynamic programming \cite{Xiao2014} and ant colony optimization \cite{ttp_ant} have been proposed to directly route the vehicles in the dynamic network. More recently, deep reinforcement learning has also been proposed for end-to-end routing without individual road segment travel time prediction \cite{Panov2018,Koh2020,Geng2020}. Moreover, graph convolution networks have been proposed to embed the structure of the road network and exploit together with reinforcement learning for routing in large dynamic networks \cite{LargeDG}.

\subsection{Problem formulation}
We address the dynamic vehicle routing problem in urban networks, which seeks to minimize the overall travel time of a vehicle fleet while adapting to real-time traffic conditions. This problem, while extensively studied in transportation research, presents unique challenges when approached through multi-agent reinforcement learning due to the need for (i) real-time adaptation to dynamic traffic conditions and (ii) coordinated decision-making to avoid system-wide congestion cascades. Our formulation differs from classical shortest path routing in its emphasis on multi-agent coordination, and from fleet management problems \cite{lin2018efficient, holler2019deep, zhang2020dynamic} in its focus on individual vehicle routing rather than supply-demand balancing. The problem shares conceptual similarities with packet routing in IP networks, where discrete entities (vehicles/packets) are routed through intermediate nodes (intersections/routers) toward destinations. However, vehicular networks present distinct challenges including routing restrictions (one-way streets, turn prohibitions), physical capacity constraints, and the absence of hierarchical addressing schemes that characterize IP networks. These differences necessitate specialized modeling approaches that account for the unique spatiotemporal dynamics of urban traffic systems.

\subsection{Our approach}
To address the {\em vehicle navigation problem}, we propose a {\em network-aware multi-agent reinforcement learning (MARL)} approach with two distinct paradigms. Unlike existing MARL approaches in traffic management that focus primarily on signal control or fleet assignment, our method directly addresses individual vehicle routing through explicit coordination mechanisms. Our first contribution is the \textsc{Adaptive Navigation} (AN), a fully decentralized system that assigns a reinforcement learning agent to every intersection. When a vehicle approaches an intersection, it submits a routing query to the agent, including its final destination. The agent then generates a routing response based on the vehicle's destination and the current state of local traffic, leveraging Graph Attention Networks (GAT) for neighbor communication and emergent coordination.

\smallskip\noindent However, assigning an agent to every intersection is not feasible for large, real-world city networks due to scalability constraints. To overcome this challenge, we introduce {\em \textsc{Hierarchical Hub-based Adaptive Navigation} (HHAN)}, a hierarchical and scalable extension of our model. In this framework, agents are placed only at a strategically selected subset of critical intersections, referred to as {\em hubs}. A vehicle's journey is thus decomposed into a sequence of long-range, hub-to-hub navigations. To enhance coordination among hub agents in HHAN, we employ the {\em Attentive Q-Mixing (A-QMIX)} framework following a centralized training with decentralized execution paradigm. Building upon the QMIX value decomposition approach \cite{rashid2018qmixmonotonicvaluefunction}, our method extends it to handle asynchronous agent decisions through attention-based aggregation. This approach allows agents to learn a shared, global value function during training, enabling them to discover collaborative routing policies that minimize system-wide congestion while maintaining decentralized execution. The feasibility of our approach relies on {\em vehicle-to-infrastructure (V2I)} communication, where vehicles query hub agents for their next routing directive. This is enabled by technologies such as DSRC \cite{khan2022dsrc}, and can be applied to both conventional and autonomous vehicles \cite{2wrd_cnnctd_AVs,Chen2020}.

\subsection{Contributions}
Our work makes the following key technical contributions to multi-agent reinforcement learning for dynamic vehicle routing:
\begin{itemize}
    \item \textbf{Adaptive Navigation (AN)}. We develop {\em \textsc{Adaptive Navigation} (AN)}, a fully decentralized MARL approach for coordinated vehicle routing, incorporating Graph Attention Networks for intersection-level coordination and emergent multi-agent behavior.
    \item \textbf{Hierarchical Hub-based Adaptive Navigation (HHAN)}. We introduce {\em \textsc{Hierarchical Hub-based Adaptive Navigation} (HHAN)}, a scalable hierarchical extension of the \textsc{Adaptive Navigation} model using strategically-selected {\em hub agents}. HHAN employs the {\em Attentive Q-Mixing (A-QMIX)} framework for centralized training with decentralized execution to address large-scale networks. A-QMIX is a coordination mechanism that extends traditional QMIX to handle asynchronous multi-agent decisions through a novel {\em attention-based aggregation} method operating over {\em Global Collection Epochs (GCE)}.
    \item \textbf{Spatial locality preservation through Z-order curve encoding}. We adapt the {\em Z-order space-filling curve} for destination representation in traffic networks, providing a scalable method for preserving spatial locality while maintaining neural network separability.
    \item \textbf{Comprehensive Evaluation}. We conduct empirical evaluation on synthetic and realistic road networks, demonstrating performance improvements over established routing baselines and analyzing the learned coordination behaviors.
    \item \textbf{Open-source code}. We ensure reproducibility by making source code publicly available.
    \begin{tcolorbox}[colback=black!5,colframe=black!70,boxrule=.2mm,boxsep=.2mm,top=0.2mm]
    \smallskip\noindent \textbf{GitHub repository:}\\
    \url{https://github.com/Arianhgh/HHAN}
    \end{tcolorbox}
\end{itemize}
\subsection{Organization}
Section \ref{chap:problem} provides preliminaries and formally defines the adaptive navigation problem. Section \ref{chap:methodology} presents the methodology for both the AN model and its hierarchical extension HHAN. Section \ref{sec:evals} discusses our experimental evaluation. The related work is discussed in Section ~\ref{chap:related}. We conclude in Section \ref{chap:conclusion}.
\section{Problem Definition}
\label{chap:problem}
We define the adaptive navigation problem based on an intersection-level formulation of the road network.

\subsection{Network Model}
We model the road network as a directed graph $W = (I, R)$, where $I = \{i_1, \ldots, i_N\}$ is the set of vertices representing intersections, and $R = \{r_1, \ldots, r_M\}$ is the set of directed edges representing roads. Each road $r \in R$ is a directed edge from $r.head \in I$ to $r.tail \in I$, meaning that every road connects exactly two intersections. Let $VCs = \{vc_1, \ldots, vc_L\}$ denote the set of $L$ vehicles, and $U = \{u_1, \ldots, u_N\}$ the set of $N$ router agents, each assigned to one intersection. When a vehicle approaches an intersection, it issues a routing query to the corresponding router agent.

\begin{definition}[\textit{$q$: routing query}]
\[
q = (t, vc, u, r_c, i_d, t_{\max})
\]
A query generated at time $t$ by vehicle $vc$ (where $vc$ denotes the vehicle identifier) currently traveling on road $r_c$, addressed to router agent $u$ assigned to intersection $r_c.tail$.\footnote{We use dot notation to denote object attributes.} The query specifies the destination intersection $i_d$ and the arrival deadline $t_{\max}$.
\end{definition}

\smallskip\noindent The router $q.u$, located at the tail of the vehicle's current road, responds to $q$ with a routing decision. To define this, we first introduce the next-hop road set.

\begin{definition}[\textit{$NH(r)$: next-hop road set}]
\[
NH(r) = \{ r_k \in R \;|\; r_k.head = r.tail, \ r_k \text{ is connected to } r \}
\]
The set of outgoing roads from intersection $r.tail$ that can be reached directly from $r$. A road $r_k$ is considered connected to $r$ if the intersection's traffic rules allow travel from $r$ to $r_k$ (e.g., no U-turns or prohibited turns).
\end{definition}

\begin{definition}[\textit{$resp(q)$: routing response}]
\[
resp(q) =
\begin{cases}
	\langle success \rangle,  & \text{if } q.r_c.tail = q.i_d,  \\
    \langle fail \rangle,     & \text{if } q.t_{\max} < q.t, \\
    \langle r \in NH(q.r_c) \rangle, & \text{otherwise}.
\end{cases}
\]
\end{definition}
\smallskip\noindent A routing response may indicate success, failure, or specify the next road to take.

\begin{definition}[\textit{$trip$: trip of vehicle $vc$}]
\[
trip = (vc, t, r, i, t_{\max})
\]
The trip of vehicle $vc$ starting at time $t$ from road $r$ with destination intersection $i$ and arrival deadline $t_{\max}$.
\end{definition}
\smallskip\noindent We denote the set of all trips as $Trips = \{ trip \;|\; trip.vc \in VCs \}$.

\begin{definition}[\textit{$path(trip)$: path of a trip}]
\[
path(trip) = (resp(q_1), \ldots, resp(q_z) \in \{\langle success \rangle, \langle fail \rangle\})
\]
The sequence of routing responses for the queries generated by $trip.vc$. The length of a path is $|path| = z$ and the last element is $path_{|path|} = resp(q_z)$. The set of all paths is $Paths = \{ path(trip) \;|\; trip \in Trips \}$.
\end{definition}

\begin{definition}[\textit{$tt(p)$: travel time of path $p$}]
\[
tt(p) = (q_{|p|}.t) - (q_1.t)
\]
The difference between the timestamps of the last and first queries in path $p$.
\end{definition}

\begin{definition}[\textit{$RS$: Routing Success}]
\[
RS = \{ p \in Paths \;|\; p_{|p|} = \langle success \rangle \}
\]
The set of paths that end with a $\langle success \rangle$ response.
\end{definition}

\begin{definition}[\textit{$AVTT$: average travel time}]
\[
AVTT = \frac{\sum_{p \in RS} tt(p)}{|RS|}
\]
The average travel time of all successful paths.
\end{definition}

\begin{definition}[\textit{Locality of access}]
Let $D(i,j)$ denote the Euclidean distance between intersections $i$ and $j$, and $E(T(i,j))$ the expected travel time between them. A network satisfies locality of access if, for intersections $i_1, i_2, i_3 \in I$ with $D(i_1, i_2) > D(i_1, i_3)$, it holds that $E(T(i_1, i_2)) > E(T(i_1, i_3))$. This property is crucial for efficient destination representations, as preserved in our Z-order encoding.
\end{definition}

\subsection{Adaptive Navigation Problem}

\noindent We now formally define the \textit{adaptive navigation problem}.

\begin{definition}[\textit{Adaptive Navigation Problem}]
Given a road network $W$ and a set of routing queries $Q$, the objective is to generate a routing response $resp(q)$ for each $q \in Q$ so as to:
\begin{enumerate}
    \item maximize $|RS|$, the number of successful routes, and
    \item minimize $AVTT$, the average travel time over all successful routes.
\end{enumerate}
\end{definition}

\section{Methodology}
\label{chap:methodology}

We present our methodology in two parts. The first part describes the \textsc{Adaptive Navigation} (AN), where an agent is assigned to each intersection in a fully decentralized multi-agent reinforcement learning (MARL) approach. The second part introduces a scalable extension that employs a hierarchical hub-based structure with centralized training and decentralized execution to address large-scale networks effectively.

\medskip\noindent\textbf{Multi-Agent Paradigm Overview}. Before detailing our methodology, we clarify the multi-agent paradigms employed. In \textbf{decentralized systems}, agents operate independently with local observations and decision-making, relying on emergent coordination through shared environment dynamics. Our AN model exemplifies this approach, with intersection agents making autonomous routing decisions based on local traffic states and limited neighborhood information via Graph Attention Networks. In contrast, \textbf{centralized training with decentralized execution (CTDE)} systems train agents using global information but deploy them with only local observations for scalability. Our HHAN model follows this paradigm, using centralized coordination during training through the A-QMIX framework while maintaining decentralized execution capabilities. The choice of paradigm reflects the inherent trade-offs between coordination effectiveness and computational scalability in multi-agent traffic systems.

\subsection{Adaptive Navigation (AN)}
\label{subsec:intersection_model}

In our \textsc{Adaptive Navigation} (AN) model, we formulate the traffic routing problem as a decentralized MARL task, assigning a unique agent to each intersection to handle routing decisions. This fully distributed approach leverages local information while achieving implicit coordination through shared network states and Q-learning updates. Below, we detail the formulation, state representations, actions, rewards, and learning process.

\subsubsection{MARL Formulation}

\noindent\textbf{Router agent at intersection $i$, $u_i$.} We assign a unique agent $u_i$ to each intersection $i \in I$. The agent $u_i$ responds only to queries $q \in Q$ where the tail of the current road segment $q.r_c.tail$ equals $i$. This ensures that each agent focuses on decisions relevant to its specific intersection.

\noindent\textbf{State of query $q$, $s_q$.} The state of a query $q$ is defined as the unique representation of its destination intersection, denoted as $[q{\text -}i_d]$. We discuss efficient representations for destination IDs in Section \ref{subsec:Intersec ID}:
\[
s_q = [q{\text -}i_d]
\]

\noindent\textbf{State of intersection $i$ at time $t$, $s_i^t$.} The state of intersection $i$ at time $t$ captures the congestion status of its outgoing roads. A road $r \in R$ with $r{\text -}head == i$ is considered congested ($C(r) = True$) if its current speed is below a fixed proportion (defined by the hyperparameter \texttt{congestion-speed-factor}) of its free-flow speed. The state $s_i^t$ is zero-extended to a fixed dimension $\mathbb{R}^F$, where $F$ is the maximum number of outgoing roads among all intersections:
\[
s_i^t = [[1 \text{ if } C(r) == True \text{ else } 0]] \quad \forall r \in R, r{\text -}head == i
\]

\noindent\textbf{State of road network $W$ at time $t$, $s_W^t$.} The network state at time $t$ is the concatenation of all intersection states:
\[
s_W^t = [s_1^t | \dots | s_N^t]
\]

\noindent\textbf{State of query $q$ at step $\tau$, $s_q^\tau$.} For a query $q$ associated with vehicle $q{\text -}vc$ at step $\tau$ in its path, the state is a tuple combining the query's destination representation and the network state at the query's time:
\[
s_q^\tau = (s_q, s_W^{q{\text -}t})
\]
\label{eqn:s_q}

\noindent\textbf{Action of agent $u_i$ for $s_q^\tau$, $a(s_q^\tau)$.} The action is the selection of an outgoing road segment from the intersection $i = q{\text -}r_c{\text -}tail$:
\[
a(s_q^\tau) = resp(q)
\]

\noindent\textbf{Next state of $s_q^\tau$, $s_q^{\tau+1}$.} For the next query $q^\prime$ in the vehicle's path (i.e., at step $\tau+1$):
\[
s_q^{\tau+1} = (s_{q^\prime}, s_W^{q{\text -}t^\prime})
\]

\noindent\textbf{Reward, $r(a(s_q^\tau))$.} The reward is defined as the negative time difference between consecutive queries in the vehicle's path:
\[
\Delta T = (q^\prime{\text -}t) - (q{\text -}t)
\]
\[
r(a(s_q^\tau)) = -\Delta T
\]
\label{eqn:reward}
The justification for this reward function, which incentivizes minimizing travel time, is discussed in Section \ref{sec:rewardfunc}.

\noindent\textbf{Network State Aggregation with GAT.} To provide agents with relevant traffic context, we aggregate the network state using a Graph Attention Network (GAT). The GAT takes the network state $s_W^t = \{s_1^t, \dots, s_N^t\}$, where $s_i^t \in \mathbb{R}^F$, as input and produces a local embedding $s_i^\prime \in \mathbb{R}^{F^\prime}$ for each agent $u_i$. A shared linear transformation, parameterized by a weight matrix $\omega \in \mathbb{R}^{F^\prime \times F}$, transforms the input features into higher-level features:
\[
H_i = \omega \cdot s_i^t
\]
A self-attention mechanism computes attention coefficients $e_{ij} = a(H_i, H_j)$ for nodes $j \in N_i$ (the one-hop neighborhood of $i$, including $i$ itself). These coefficients are normalized using a softmax function:
\[
\alpha_{ij} = \text{softmax}(e_{ij}) = \frac{\exp(e_{ij})}{\sum_{k \in N_i} \exp(e_{ik})}
\]
The GAT output for intersection $i$ is:
\[
(GAT(s_W^t))_i = s_i^\prime = \sigma\left(\sum_{j \in N_i} \alpha_{ij} H_j\right)
\]
\label{eqn:att-layer}
Multiple attention heads stabilize the learning process, with concatenation for intermediate layers and averaging for the final layer. The number of GAT layers is a tunable hyperparameter, controlling the extent of neighborhood information aggregation.

\noindent\textbf{State Representation for Learning.} The destination $s_q$ is processed through a linear layer with ReLU activation to produce embeddings:
\[
[s_q] = \text{ReLU}(\text{Linear}_i(s_q))
\]
The routing response is the action with the highest Q-value:
\[
resp(q) = \argmax_a Q_i(s_q^\tau, a)
\]
Each agent uses a Q-learning algorithm with the MSE loss:
\[
L(s_q^\tau, a, r: \theta) = \mathbb{E}\left[\left(r + \gamma \max_{a^\prime} Q_{i+1}(s_q^{\tau+1}, a^\prime) - Q_i(s_q^\tau, a)\right)^2\right]
\]
\label{eqn:loss_iql}
The intertwined Q-learning updates enable implicit coordination, as the value of a state at agent $u_i$ depends on the next agent's Q-values.

\subsubsection{Destination Representation}
\label{subsec:Intersec ID}

The destination representation must be unique, separable by the neural network, low-dimensional, and preserve locality of access. We compare two approaches and propose the Z-order curve for optimal performance.

\begin{figure}[t]
    \centering
    \begin{minipage}[t]{0.4\linewidth}
        \includegraphics[width=\linewidth]{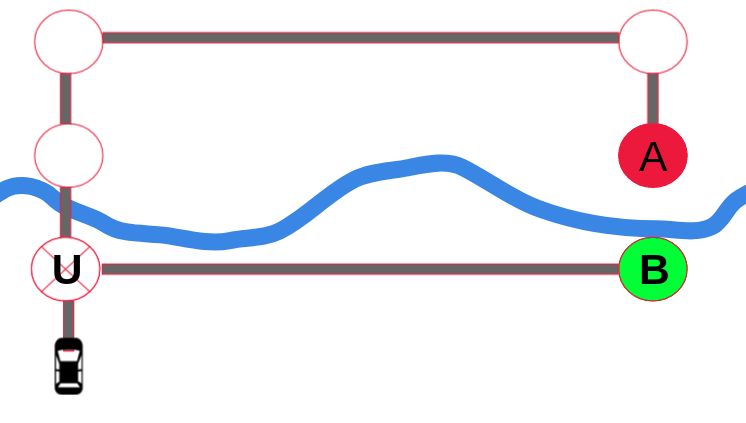}
        \caption{Hard to separate destinations.}
        \label{fig:cord-prblm}
    \end{minipage} \qquad
    \begin{minipage}[t]{0.4\linewidth}
        \includegraphics[width=0.9\linewidth]{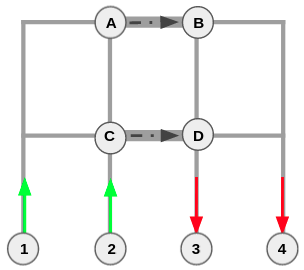}
        \caption{Collaborative policies.}
        \label{fig:colab_pol}
    \end{minipage}
    \vspace{-15pt}
\end{figure}

\noindent\textbf{Coordinates and One-Hot IDs.} Normalized intersection coordinates are unique, low-dimensional (dimension = 2), and preserve locality but are hard for neural networks to separate for nearby intersections (see Figure \ref{fig:cord-prblm}). One-hot encodings are unique and separable but high-dimensional (dimension = $N$) and discard locality information.

\begin{figure}[t]
    \centering
    \begin{subfigure}[t]{0.49\linewidth}
        \centering
        \includegraphics[width=0.8\linewidth]{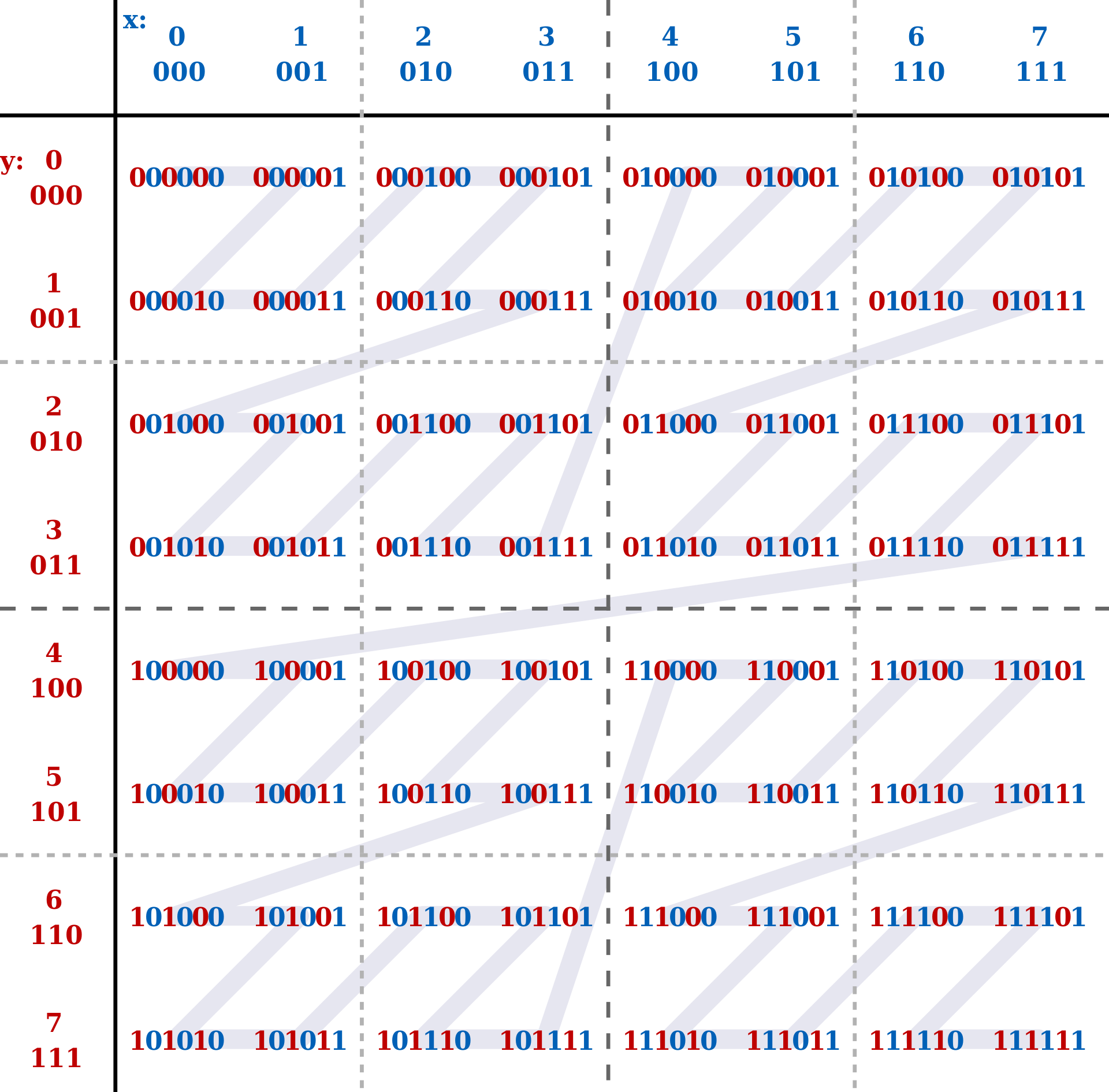}
        \caption{Interleaving the coordinates \cite{z-order}.}
        \label{fig:zordercord}
    \end{subfigure}\hfill
    \begin{subfigure}[t]{0.49\linewidth}
        \centering
        \includegraphics[width=0.8\linewidth]{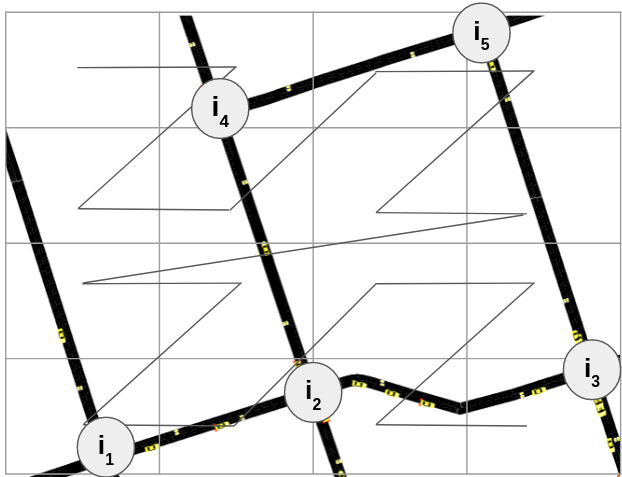}
        \caption{Depth-first traversal on quad-tree.}
        \label{fig:zorderintersec}
    \end{subfigure}
    \vspace{-5pt}
    \caption{Z-order curve (Morton Space Filtering).}
    \vspace{-15pt}
\end{figure}

\noindent\textbf{Z-order ID.} We propose using the Z-order curve \cite{morton1966computer} to create unique, linearly separable intersection IDs that partially preserve locality while maintaining a low dimension ($\log_2(N)$). The Z-order curve interleaves the binary representations of a point’s coordinates to compute a Z-value, sorting points accordingly. This is equivalent to a depth-first traversal of a quad-tree, forming Z-shapes (see Figures \ref{fig:zordercord} and \ref{fig:zorderintersec}). For an intersection $i_1$ with Z-order index 2, its ID is $\text{binary}(2) = [0,1,0]$.

\subsubsection{Algorithm Sketch}
\label{sec:Alg}

Algorithm \ref{alg:adaptive_routing} outlines the inference process at time step $t$, taking the network state $s_W^t$ and queries $Q$ as inputs to generate routing responses. Algorithm \ref{alg:training} describes the training process for the agents.

\begin{algorithm}[t]
    \begin{algorithmic}[1]
        \REQUIRE {
            state of the road network $s_W^t$,
            set of all the routing queries $Q$ at time $t$
        }
        \ENSURE optimal $resp(q)$ for $q \in Q$
        \vspace{10pt}
        \FORALL{$q \in Q$}
            \STATE $s_q \gets$ state of query $q$
            \STATE $u \gets q{\text -}u$, the router agent
            \STATE $u.memory$.push(\text{previous experience tuple of agent $u$})
            \STATE $[s_q] \gets \text{ReLU}(\text{Linear}_u(s_q))$
            \STATE $s_i^\prime \gets GAT((s_W^t))[i]$
            \STATE $s_{agg} \gets s_i^\prime$
            \STATE $r_q \gets \argmax Q\text{-}net_u([[s_q]|s_{agg}])$
        \ENDFOR
        \IF{in training mode}
            \STATE Train($\{q{\text -}u | q \in Q\}$)
        \ENDIF
    \end{algorithmic}
    \caption{Inference at time $t$}
    \label{alg:adaptive_routing}
\end{algorithm}

\begin{algorithm}[t]
    \begin{algorithmic}[1]
        \REQUIRE {
            set of router agents $U$ that need training
        }
        \ENSURE training of RL agents
        \vspace{10pt}
        \FORALL{$u \in U$}
            \IF{time-to-learn(u)}
                \STATE training-batch = u.memory.sample()
                \STATE loss = MSE-loss(training-batch)
                \STATE loss.backward()
                \STATE UpdateGATParameters()
            \ENDIF
        \ENDFOR
    \end{algorithmic}
    \caption{Train}
    \label{alg:training}
\end{algorithm}

\subsection{Scalable Extension: Hierarchical Hub-based Adaptive Navigation (HHAN)}

To enhance scalability for large-scale road networks, we propose the \textsc{Hierarchical Hub-based Adaptive Navigation} (HHAN) model. Instead of placing agents at every intersection, we strategically select a subset of key intersections (hubs) and assign agents to them. These agents coordinate through a centralized training scheme with the Attentive Q-Mixing (A-QMIX) framework, enabling efficient routing decisions across expansive networks.

\subsubsection{Hierarchical Hub Abstraction}

In large-scale, real-world road networks, assigning and coordinating agents at every intersection can be computationally expensive and operationally impractical. To address this, HHAN introduces a \textit{hierarchical hub abstraction} that reduces the complexity of decision-making by focusing only on a strategically selected subset of intersections. This abstraction is built directly on top of the underlying road network graph $W$.

\begin{definition}[\textit{Hub Network}]
The road network $W$ is abstracted into a directed \textit{hub graph} $W_H = (H, E_H)$, where $H = \{h_1, \ldots, h_K\}$ is a set of $K$ hubs, with $K \ll N$. Each hub $h_k \in H$ corresponds to a strategically chosen intersection from $I$ based on criteria such as high traffic centrality, network connectivity, or bottleneck potential. An edge $(h_a, h_b) \in E_H$ exists if there is at least one viable route from hub $h_a$ to hub $h_b$ in the original road network $W$.
\end{definition}

\smallskip\noindent In this abstraction, the vehicle navigation problem shifts from making local routing decisions at every intersection to making higher-level strategic decisions only at hubs. Each vehicle's journey is decomposed into a sequence of \textit{hub-to-hub} segments. The micro-level routing between two hubs is delegated to a conventional Shortest Path First (SPF) algorithm, which ensures efficiency in low-level navigation while the hub-level agents focus on global coordination and congestion management.

\begin{definition}[\textit{Hub-level routing query}]
When a vehicle arrives at a hub $h_k$, it submits a routing query to the corresponding hub agent $u_k$. The query contains the vehicle's current hub and its final destination. Instead of returning a single next road segment, the hub agent selects the next hub $h_{\text{next}} \in H$ to navigate towards, based on its learned policy and current traffic conditions.
\end{definition}

\smallskip\noindent This hierarchical approach provides two major advantages. First, it significantly reduces the number of agents, making the problem tractable for metropolitan-scale networks. Second, by operating at a higher level of abstraction, hub agents can coordinate more effectively to prevent downstream congestion, rather than reacting only to immediate local traffic conditions. In HHAN, the hub network serves as a strategic decision layer, while standard SPF routing ensures fine-grained vehicle movement between hubs. This design balances scalability, adaptability, and coordination in a unified framework.

\subsubsection{Hub Selection and Connectivity}

The effectiveness of HHAN relies on the careful selection and connectivity of hubs, which serve as critical decision points in the network.

\begin{enumerate}
    \item \textbf{Candidate Filtering:} We identify significant intersections by selecting nodes with an in-degree of at least three and an out-degree of at least three, ensuring that these nodes correspond to major junctions with sufficient directional connectivity to meaningfully affect routing decisions.
    \item \textbf{Hub Selection:} From these candidates, we use the K-Medoids clustering algorithm with shortest-path distance as the metric to select hubs. This approach identifies the most central intersection (medoid) within each cluster, ensuring both centrality and spatial distribution across the network.
    \item \textbf{Hub Connectivity:} To form the hub graph $W_H$, each hub is connected to at most $k = 3$ nearest neighboring hubs based on shortest-path travel time. A connection is established only if the Euclidean distance between hubs is below a threshold $d_{\text{max}}$, where $d_{\text{max}}$ is chosen according to the scale of the map to prevent impractical long-distance routing.

\end{enumerate}

\smallskip\noindent This structured approach ensures that hubs are strategically placed to cover the network efficiently while maintaining feasible routing paths.

\subsubsection{Hierarchical Agent Formulation}

In this model, an agent $u_k$ is assigned to each hub $h_k \in H$. Unlike the foundational model, where agents act at specific intersections, hub agents are triggered when a vehicle enters their operational vicinity, defined as a radius $r_{\text{vic}}$ around the hub. The value of $r_{\text{vic}}$ is chosen according to the scale of the map and is set as half of the minimum distance between two neighboring hubs. This vicinity-based approach provides navigational flexibility, allowing agents to make decisions based on broader traffic patterns. The agent’s action is to select the next hub $h_{\text{next}}$ for the vehicle to travel toward. If $h_{\text{next}}$ is the vehicle’s final destination hub, the vehicle is routed directly to its final road edge using a standard Shortest Path First (SPF) algorithm.

\subsubsection{Flow-Aware State Representation}

To enable agents to make informed decisions, we design a state representation that captures both local and predictive traffic flow dynamics, replacing the GAT with a fixed-size representation.

\begin{definition}\textit{Local Observation $\tau_k$}

The local observation for agent $u_k$ is a concatenated vector:
\[
\tau_k = \text{concat}(Emb(h_d), F_k, F_{N(k)})
\]
\begin{enumerate}
    \item \textit{Destination Hub Embedding $Emb(h_d)$:} The Z-order embedding of the vehicle’s destination hub $h_d$, as described in Section \ref{subsec:Intersec ID}, ensuring a unique and locality-preserving representation.
    \item \textit{Current Hub Features $F_k$:} A feature vector for hub $h_k$, including:
        \begin{itemize}
            \item \textbf{Vicinity Speed:} The average normalized speed of vehicles within a radius $r_{\text{vic}}$ of the hub. This radius captures approaching and departing traffic, providing predictive insights into potential congestion.
            \item \textbf{Outgoing Congestion Ratio:} The average ratio of current travel time to free-flow travel time on edges within a radius $r_{\text{vic}}$ of the hub , indicating the ease of traffic dispersal from the hub.
        \end{itemize}
    \item \textit{Padded Neighbor Features $F_{N(k)}$:} A fixed-size feature vector for up to $M_{\text{neighbors}}$ neighboring hubs, each containing:
        \begin{itemize}
            \item \textbf{Estimated Travel Time:} A normalized estimate of travel time from $h_k$ to neighbor $h_j$, reflecting real-time conditions.
            \item \textbf{Neighbor Congestion Ratio:} The average congestion ratio on edges within a radius $r_{\text{vic}}$ of the hub $h_j$, providing information about downstream conditions.
            \item \textbf{Distance to Destination:} The normalized static network distance from $h_j$ to the destination hub $h_d$, aiding in long-term routing decisions.
        \end{itemize}
\end{enumerate}
\end{definition}

\smallskip\noindent This flow-aware representation equips agents with an understanding of traffic dynamics, enabling proactive routing decisions.

\subsubsection{Coordinated Training with Attentive Q-Mixing (A-QMIX)}

To achieve robust coordination in an asynchronous system, we adopt a Centralized Training with Decentralized Execution (CTDE) paradigm based on the QMIX framework \cite{rashid2018qmixmonotonicvaluefunction}, introducing the Global Collection Epoch (GCE) to bundle decisions over time.

\noindent Global Collection Epoch (GCE) A GCE aggregates all decisions made across the system within a time period or a fixed number of decisions into a transition tuple 
\[
(\mathbf{s}, \{\mathcal{D}_k\}_{k=1}^K, r, \mathbf{s}'),
\] 
where $\mathbf{s}$ is the global state, $\{\mathcal{D}_k\}_{k=1}^K$ is the set of decisions made by all agents $u_1, \dots, u_K$, $r$ is the aggregated reward, and $\mathbf{s}'$ is the resulting global state after executing all the decisions in the GCE.

\begin{definition}\textit{Global State $\mathbf{s}$}

The global state captures system-wide flow properties:
\begin{enumerate}
    \item \textit{All-Hub Flow Snapshot:} For each hub, the vicinity speed and edge congestion ratio within operational vicinity, providing a network-wide view of traffic bottlenecks.
    \item \textit{System-Wide Efficiency:} Metrics such as the total number of active vehicles, completion throughput ratio (completed vs. started trips), and average trip inefficiency (actual vs. shortest path travel time).
    \item \textit{System Imbalance:} The standard deviation of vicinity speeds across hubs, quantifying traffic flow imbalance.
\end{enumerate}
\end{definition}

\noindent\textbf{Attentive Q-Mixing (A-QMIX).} In this framework, each agent $u_k$ maintains a local Q-network 
\[
Q_k(\tau_k, a_k),
\] 
which estimates the expected cumulative reward for taking action $a_k$ given the local observation $\tau_k$. Unlike standard MARL settings where agents act synchronously, in our traffic control environment, agents make multiple decisions asynchronously within a Global Collection Epoch (GCE). To handle this asynchrony, we introduce an \textit{attention mechanism} that aggregates an agent's multiple local decisions into a single utility score. 

\smallskip\noindent For agent $u_k$ with decision set $\mathcal{D}_k = \{d_1, d_2, \ldots, d_{|\mathcal{D}_k|}\}$ in the current GCE, we compute attention weights $\alpha_{k,i}$ for each decision $d_i \in \mathcal{D}_k$ as:
\[
\alpha_{k,i} = \frac{\exp(\mathbf{w}^\top \tanh(\mathbf{W}_1 [\mathbf{s}; \tau_{k,d_i}; Q_k(\tau_{k,d_i}, a_{k,d_i})]))}{\sum_{j=1}^{|\mathcal{D}_k|} \exp(\mathbf{w}^\top \tanh(\mathbf{W}_1 [\mathbf{s}; \tau_{k,d_j}; Q_k(\tau_{k,d_j}, a_{k,d_j})]))},
\]
where $\mathbf{W}_1 \in \mathbb{R}^{h \times (|\mathbf{s}| + |\tau_k| + 1)}$ and $\mathbf{w} \in \mathbb{R}^h$ are learnable parameters, $[\cdot; \cdot]$ denotes concatenation, and $h$ is the hidden dimension. The aggregated Q-value is then computed as:
\[
Q^*_k = \sum_{i=1}^{|\mathcal{D}_k|} \alpha_{k,i} \cdot Q_k(\tau_{k,d_i}, a_{k,d_i}).
\]
This attention mechanism dynamically weighs each decision according to its relevance to the global state $\mathbf{s}$ and local context $\tau_{k,d_i}$. Critical decisions at congested hubs or along high-priority routes receive higher attention weights $\alpha_{k,i}$, amplifying their influence on the aggregated utility $Q^*_k$, while decisions in low-impact scenarios are down-weighted, reducing noise in the learning signal.  

\smallskip\noindent The aggregated utilities $Q^*_k$ from all agents are then passed to a \textit{mixing network} that produces the global value function:
\[
Q_{tot}(\mathbf{s}, \mathbf{a}; \theta),
\]  
where $\mathbf{a} = \{a_1, a_2, \dots, a_K\}$ represents the joint actions of all agents. A key property of the mixing network is the \textit{monotonicity constraint}:  
\[
\frac{\partial Q_{tot}}{\partial Q_k} \ge 0, \quad \forall k,
\]  
which guarantees that improving an individual agent's Q-value cannot decrease the global Q-value. This monotonicity enables decentralized execution: agents can greedily select actions based on their local Q-values while still optimizing the system-wide objective.  

\smallskip\noindent Training of A-QMIX is performed end-to-end using temporal-difference (TD) learning. The loss function is defined as:  
\[
L(\theta) = \mathbb{E} \left[ (y^{tot} - Q_{tot}(\mathbf{s}, \mathbf{a}; \theta))^2 \right],
\]  
where the TD target is  
\[
y^{tot} = r + \gamma \max_{\mathbf{a}'} Q_{tot}(\mathbf{s}', \mathbf{a}'; \theta^{-}).
\]  
Here, $r$ is the aggregated reward for the GCE, $\mathbf{s}'$ is the next global state, $\gamma$ is the discount factor, and $\theta^{-}$ represents the parameters of a target network, which is periodically updated to stabilize training. The use of a global reward and state ensures that agents are incentivized to learn collaborative policies that improve overall traffic flow rather than only optimizing local metrics.  

The attention-based aggregation in A-QMIX provides several benefits:  
\begin{itemize}
    \item \textbf{Handling Asynchrony:} Agents can make multiple decisions at different times, yet their contributions are combined meaningfully.  
    \item \textbf{Prioritization of High-Impact Decisions:} Critical decisions affecting congestion or bottlenecks are weighted more heavily, improving learning efficiency.  
    \item \textbf{Decentralized Execution with Global Coordination:} Monotonic mixing allows agents to act independently while still aligning with global objectives, which is crucial in real-time traffic systems.
\end{itemize}

\subsubsection{Model Architecture}

\begin{figure}[t]
    \centering
    \includegraphics[width=0.9\linewidth]{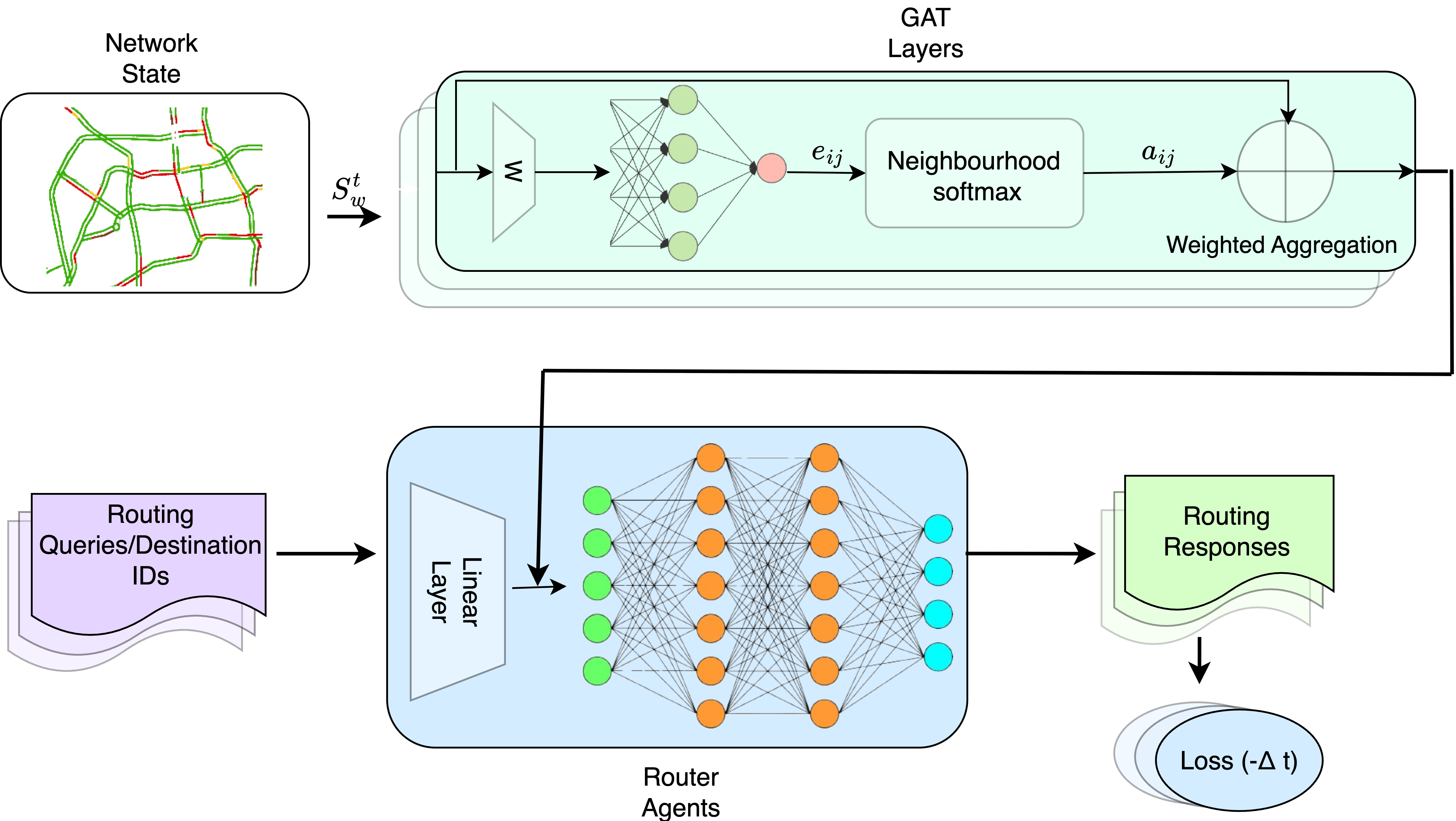}
    \caption{\textsc{Adaptive Navigation} (AN) model architecture showing the decentralized MARL approach where each intersection has an agent that processes routing queries using destination embeddings and GAT-aggregated network states. The GAT layers enable neighborhood information sharing for implicit coordination between agents, while each agent makes independent routing decisions based on local Q-networks trained with intertwined Q-learning updates.}
    \label{fig:adaptive_navigation}
    \vspace{-10pt}
\end{figure}

\begin{figure}[t]
    \centering
    \includegraphics[width=0.9\linewidth]{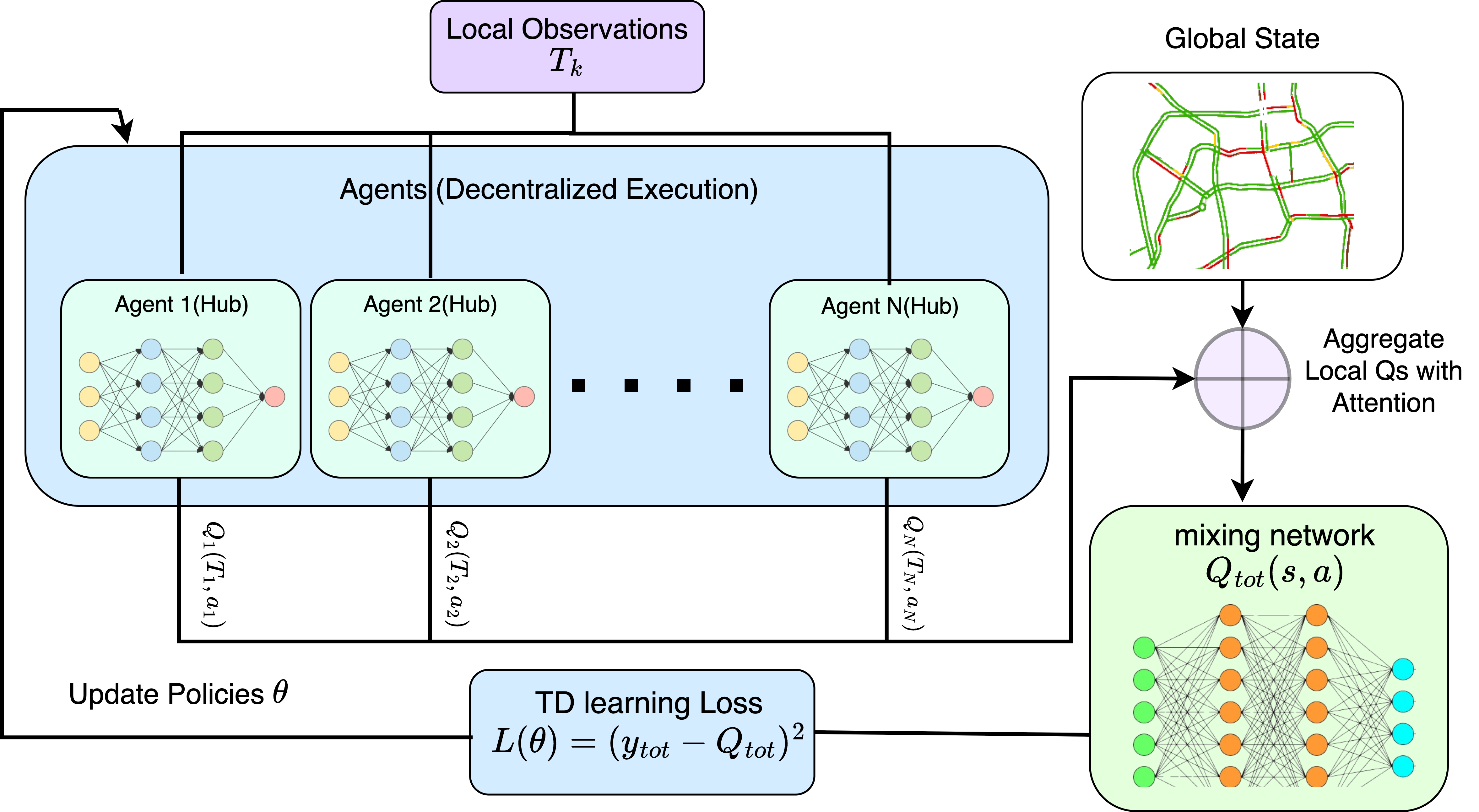}
    \caption{\textsc{Hierarchical Hub-based Adaptive Navigation} (HHAN) model architecture implementing centralized training with decentralized execution (CTDE) using the A-QMIX framework. Hub agents process local observations including destination embeddings and flow-aware features, making asynchronous routing decisions that are aggregated through an attention mechanism. The mixing network ensures monotonic value function combination while enabling coordinated learning across the hub-based network structure.}
    \label{fig:aqmix}
    \vspace{-10pt}
\end{figure}

The architecture for both approaches is depicted in Figures \ref{fig:adaptive_navigation} and \ref{fig:aqmix}. For the AN model, the destination $s_q$ is processed through a linear layer with ReLU activation, while the network state $s_W^t$ is aggregated via GAT or mean congestion methods. In HHAN, the local observation $\tau_k$ is used directly, with the centralized training framework handling coordination.

\subsection{Reward Function Justification}
\label{sec:rewardfunc}

The reward function $r(a(s_q^\tau)) = -\Delta T$ incentivizes minimizing travel time. Consider the Q-learning update rule:
\[
Q_i(s_q^\tau, a) \leftarrow Q_i(s_q^\tau, a) + \alpha \left( r_\tau + \gamma \max_{a^\prime} Q_{i+1}(s_q^{\tau+1}, a^\prime) - Q_i(s_q^\tau, a) \right)
\]
\label{eqn:rl-update}
For an infinite horizon ($\gamma=1$) and learning rate $\alpha=1$, this simplifies to:
\[
Q_i(s_q^\tau, a) = r_\tau + \max_{a^\prime} Q_{i+1}(s_q^{\tau+1}, a^\prime)
\]
\label{eqn:rl-update-simp}
Expanding for a terminal state $s_q^{\tau+Z}$:
\[
Q_i(s_q^\tau, a) = r_\tau + r_{\tau+1} + \dots + r_{\tau+Z}
\]
\label{eqn:rl-up-exp}
Substituting the reward function:
\[
Q_i(s_q^\tau, a) = -\Delta T_1 - \Delta T_2 - \dots - \Delta T_Z
\]
\label{eqn:e2e-tt}
This shows that Q-values estimate the total travel time to the destination, prioritizing states closer to the destination over faster but less direct routes.

\subsection{Justification of the MARL Architecture}

An alternative single-agent RL approach introduces high variance, as identical actions at different intersections can lead to divergent outcomes (e.g., action 0 at $i_1$ goes north, but at $i_2$ goes south). This variance hinders learning the underlying routing logic. Following \cite{Ali2020, You2020, Boyan1994, zhang2021multi}, our MARL formulation reduces variance by assigning agents to specific intersections, ensuring consistent action-state mappings. The intertwined Q-learning updates (Equation \ref{eqn:rl-update}) enable collaborative policies, as agents consider the Q-values of neighboring agents, fostering system-wide optimization.

\smallskip\noindent For example, in Figure \ref{fig:colab_pol}, the SPF algorithm may oscillate between bridges AB and CD under high load, causing congestion. The MARL approach explores collaborative policies, such as splitting traffic between bridges based on destinations, improving efficiency \cite{Boyan1994}.

\subsection{Hyper-parameter Settings}
\label{sec:hyper-param}
Tables \ref{tab:qlearn}, \ref{tab:aqmix} and \ref{tab:gat} summarize the Hyper-parameter Settings.

\begin{table}[t]
    \centering
    \caption{AN Model Hyper-parameters.}
    \label{tab:qlearn}
    \begin{tabularx}{\textwidth}{LCLC}
        \toprule
        \textbf{Parameter} & \textbf{Value} & \textbf{Parameter} & \textbf{Value} \\
        \midrule
        Optimizer & Adam & Optimizer eps & 1e-4 \\
        learning rate & 0.01 & batch-size & 64 \\
        batch-norm & False & gradient-clipping-norm & 5 \\
        buffer-size & 10000 & num-new-exp-to-learn & 1 \\
        tau & 0.01 & discount rate & 0.99 \\
        \makecell[l]{epsilon-decay-rate-\\denom} & num episodes/100 & \makecell[l]{stop-exploration-\\episode} & num-eps-10 \\
        \makecell[l]{linear-hidden-units-size \\ AN(0hop)} & [8,6] & \makecell[l]{linear-hidden-units-size \\ AN(1hop)} & [10,6] \\
        \makecell[l]{linear-hidden-units-size \\ AN(2hop)} & [12,9,6] &  &  \\
        \bottomrule
    \end{tabularx}
\end{table}

\begin{table}[t]
    \centering
    \caption{Graph Attention Network Hyper-parameters.}
    \label{tab:gat}
    \begin{tabularx}{\textwidth}{LCLC}
        \toprule
        \textbf{Parameter} & \textbf{Value} & \textbf{Parameter} & \textbf{Value} \\
        \midrule
        Optimizer & Adam & num-heads-per-layer & 3 \\
        Optimizer eps & 1e-4 & learning rate & 0.01 \\
        add-skip-connection & False & bias & True \\
        dropout & 0.6 & layer-0 output dimension & 7 \\
        intersection state dimension & 4 & layer-1 output dimension & 10 \\
        \bottomrule
    \end{tabularx}
\end{table}

\begin{table}[t]
    \centering
    \caption{HHAN Model Hyper-parameters.}
    \label{tab:aqmix}
    \begin{tabularx}{\textwidth}{LCLC}
        \toprule
        \textbf{Parameter} & \textbf{Value} & \textbf{Parameter} & \textbf{Value} \\
        \midrule
        num\_hubs & 4 & hub\_agent\_dim & 64 \\
        max\_waiting\_vehicles & 40 & z\_order\_embedding\_dim & 8 \\
        num\_episodes & 500 & max\_steps\_per\_episode & 3000 \\
        lr & 0.0005 & gamma & 0.99 \\
        epsilon\_start & 1.0 & epsilon\_end & 0.05 \\
        epsilon\_decay & 0.99 & polyak & 0.995 \\
        min\_gce\_buffer\_size & 200 & gce\_buffer\_capacity & 10000 \\
        qmix\_batch\_size & 64 & qmix\_update\_frequency\_steps & 32 \\
        mixing\_hidden\_dim & 128 & mixing\_lr & 0.0005 \\
        gce\_size & 10 & gce\_max\_sim\_time & 100 \\
        clip\_grad\_norm & 10.0 & & \\
        \bottomrule
    \end{tabularx}
\end{table}

\section{Experimental Evaluation}
\label{sec:evals}

This section provides an empirical evaluation of our proposed traffic routing models. We utilize the open-source microscopic traffic simulator, Simulation of Urban Mobility (SUMO) \citep{SUMO2018}, to create reproducible testing environments. Our evaluation examines two approaches: first, we assess the performance of the \textsc{Adaptive Navigation} (AN) model. Second, we evaluate the scalability of the \textsc{Hierarchical Hub-based Adaptive Navigation} (HHAN) model on large-scale networks. Performance is benchmarked against established routing algorithms across synthetic and realistic road networks. Our primary evaluation metrics are \textbf{Average Vehicle Travel Time (AVTT)}, which quantifies system efficiency, and \textbf{Routing Success Rate (RSR)}, defined as the percentage of vehicles that successfully reach their destination within the simulated period, measuring the system's reliability and ability to prevent gridlock \citep{wei2019survey}.

\subsection{Experimental Setup}

To ensure reproducibility, we define a experimental protocol covering the simulation environment, network topologies, traffic demand profiles, baseline algorithms, and model configurations.

\subsubsection{Simulation Environment and Metrics}
All experiments were executed using SUMO, controlled via its Python API, TraCI \citep{SUMO2018}. The simulations were run on a server equipped with 2 × Intel Xeon E5-2687W v4 3.0 GHz 12-Core Processors (30 MB L3 Cache), 512 GB of RAM (8 × 64 GB), and 8 × NVIDIA MSI GeForce GTX 1080Ti 11 GB Aero OC GPUs for accelerating neural network training.

The core performance metrics are formally defined as:
\begin{itemize}
    \item \textbf{AVTT:} $AVTT = \frac{1}{|V_{completed}|} \sum_{i \in V_{completed}} (t_{arrival, i} - t_{depart, i})$, where $V_{completed}$ is the set of vehicles that finished their trips, $t_{arrival, i}$ is the arrival time of vehicle $i$, and $t_{depart, i}$ is its departure time. Lower values are better.
    \item \textbf{RSR:} $RSR = \frac{|V_{completed}|}{|V_{total}|} \times 100\%$, where $|V_{total}|$ is the total number of vehicles introduced into the simulation. Higher values are better.
\end{itemize}
\begin{wrapfigure}{r}{0.25\linewidth}
    \centering
    \includegraphics[width=\linewidth]{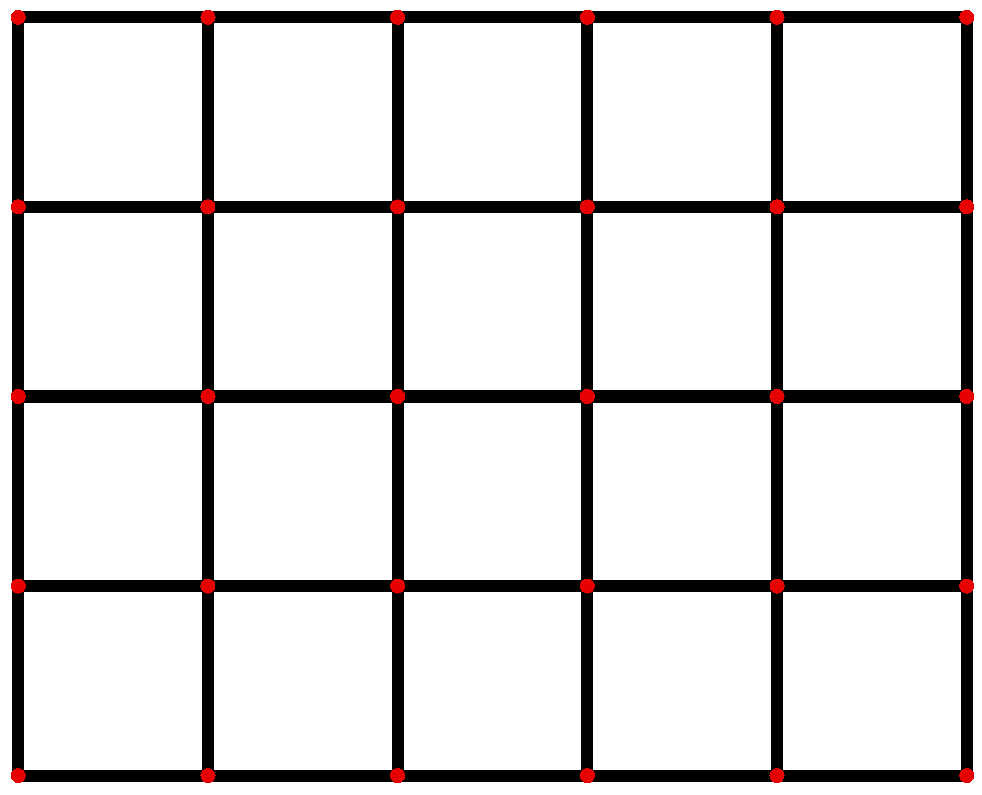}
    \caption{Synthetic 5x6 Grid road network with 30 intersections and 98 edges.}
    \label{fig:5x6}
\end{wrapfigure}
\subsubsection{Road Networks}
\begin{figure}
    \centering
    \begin{subfigure}[t]{0.48\linewidth}
        \centering
        \includegraphics[width=\linewidth]{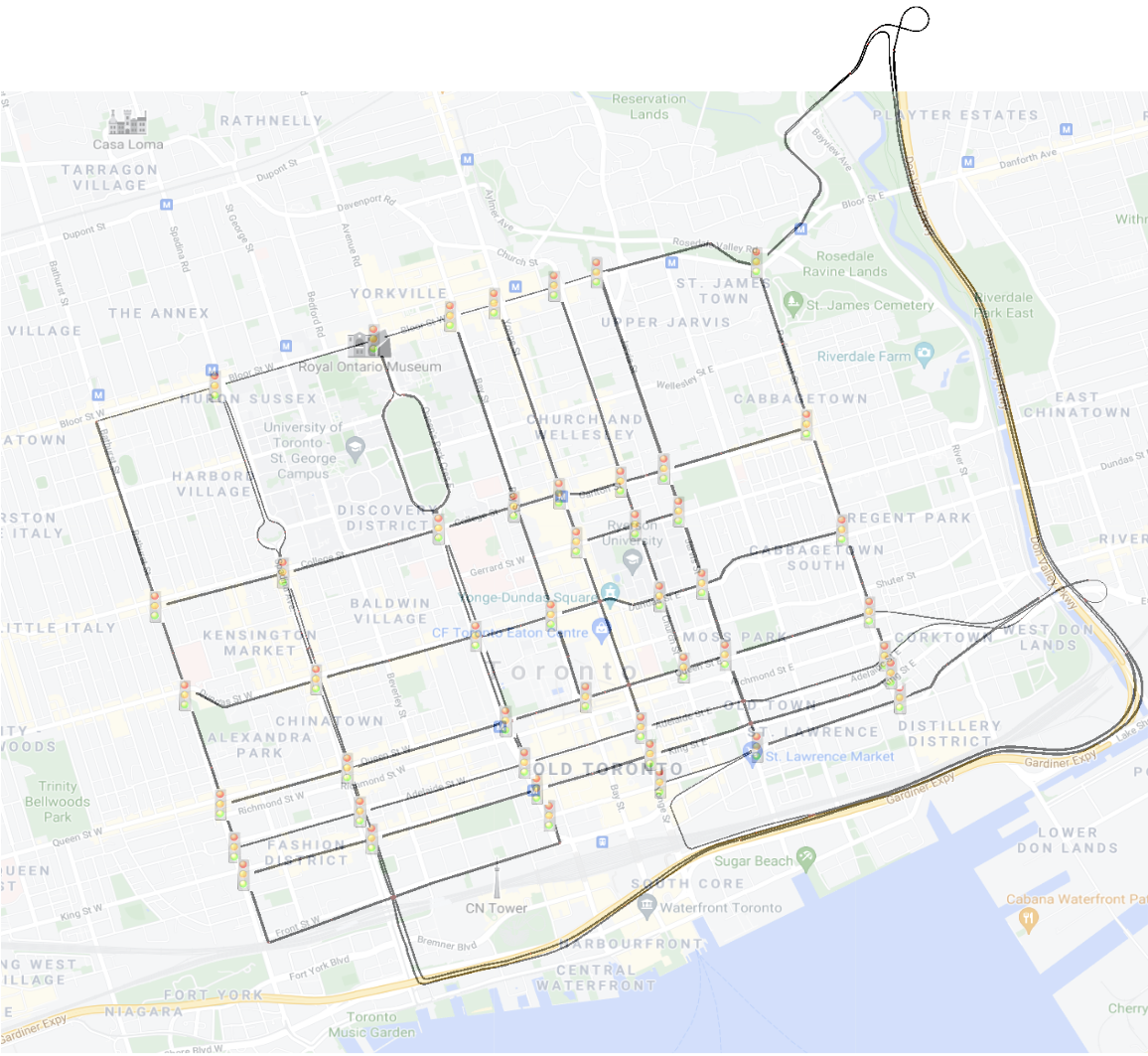}
        \caption{Downtown Toronto road network.}
        \label{fig:toronto-map}
    \end{subfigure}
    \hfill
    \begin{subfigure}[t]{0.48\linewidth}
        \centering
        \includegraphics[width=\linewidth]{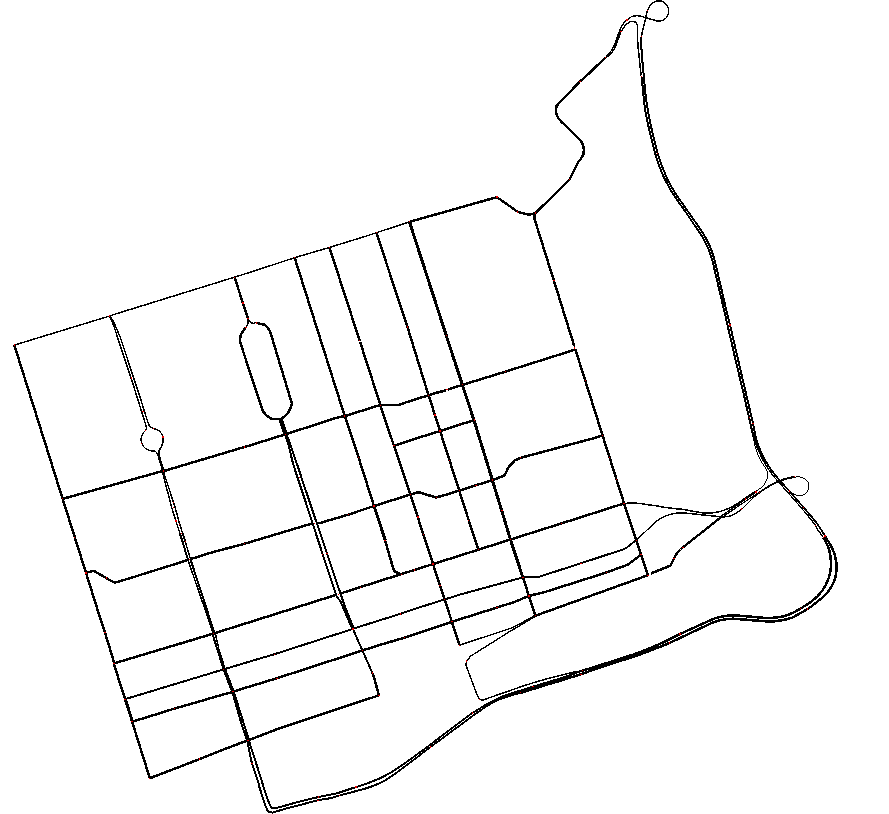}
        \caption{Abstracted Toronto network with 52 intersections and 333 edges.}
        \label{fig:toronto-overlay}
    \end{subfigure}
    \caption{Downtown Toronto road network and its abstracted version.}
    \label{fig:toronto}
\end{figure}

\begin{figure}
    \centering
    \begin{subfigure}[t]{0.48\linewidth}
        \centering
        \includegraphics[width=\linewidth]{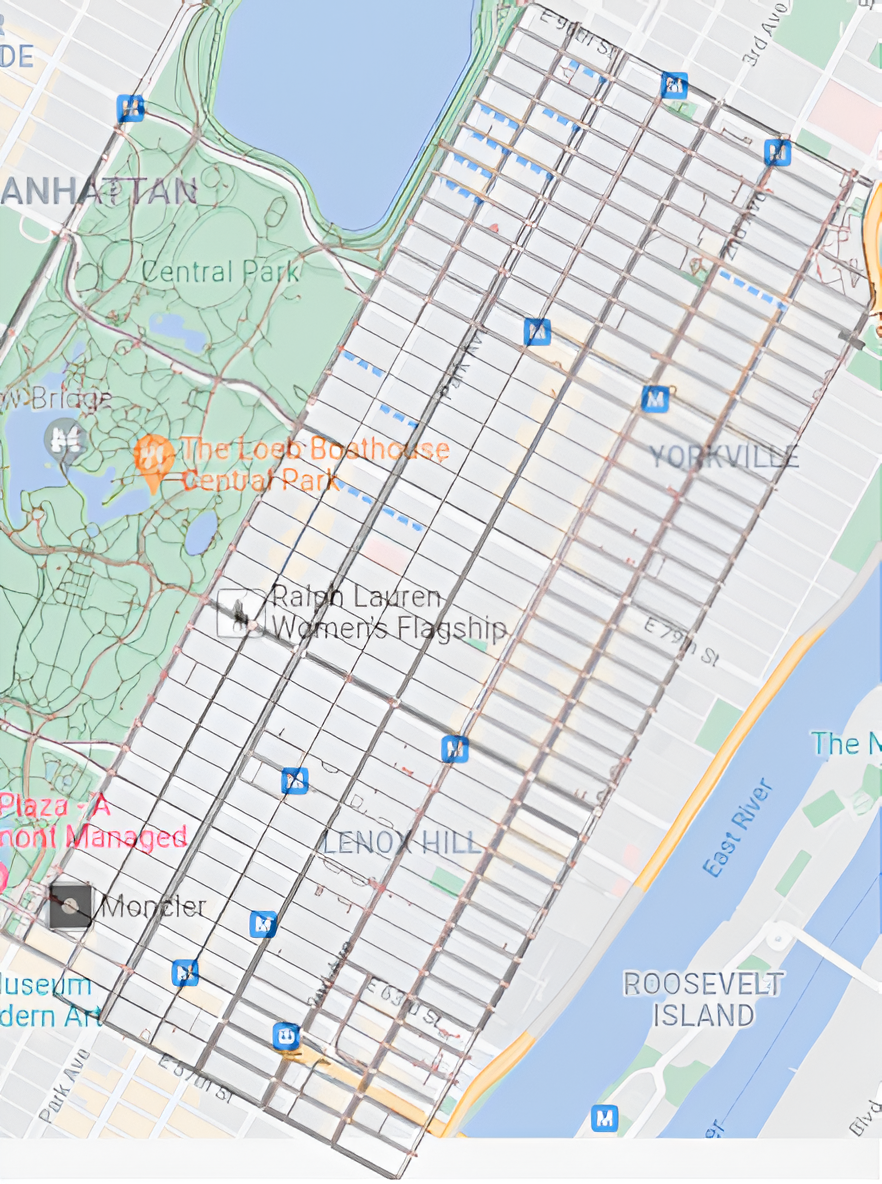}
        \caption{Manhattan road network.}
        \label{fig:manhattan-map}
    \end{subfigure}
    \hfill
    \begin{subfigure}[t]{0.48\linewidth}
        \centering
        \includegraphics[width=\linewidth]{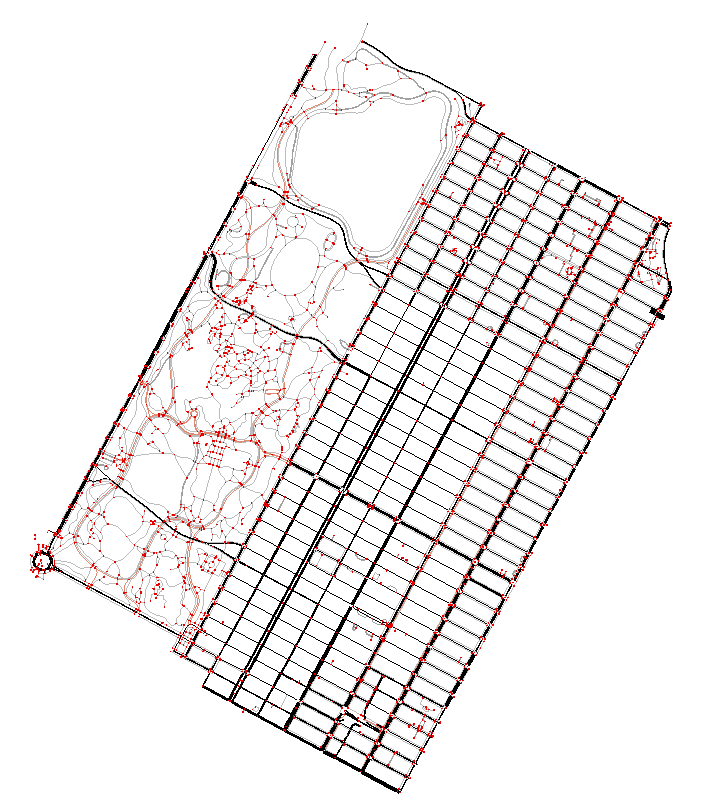}
        \caption{Abstracted Manhattan network with 320 intersections and 1184 edges.}
        \label{fig:manhattan-graph}
    \end{subfigure}
    \caption{Manhattan road network and its abstracted version.}
    \label{fig:manhattan}
\end{figure}

We employ three distinct road networks to evaluate our models under varying conditions of complexity and scale (Figures~\ref{fig:5x6}, \ref{fig:toronto}, and \ref{fig:manhattan}):

\begin{enumerate}
    \item \textbf{Synthetic 5x6 Grid:} A canonical Manhattan-style grid network consisting of 30 intersections and 98 edges (Figure~\ref{fig:5x6}). The 26 non-perimeter intersections are controlled by routing agents. All roads are two-lane with a uniform speed limit of 50 km/h. This controlled environment is ideal for analyzing model fundamentals and isolating algorithmic behaviors.
    \item \textbf{Abstracted Downtown Toronto:} A realistic network derived from OpenStreetMap data. Following the preprocessing methodology of \citep{aggarwal2021sketch,goldberg2005computing}, the raw map was simplified to 52 key intersections and 333 edges, retaining the core arterial road structure of a real-world urban center (Figure~\ref{fig:toronto}). This network features heterogeneous road lengths and speed limits, posing a more complex challenge than the synthetic grid.
    \item \textbf{Large-Scale Manhattan:} A larger network also sourced from OpenStreetMap, covering a major portion of Manhattan, NYC. It comprises 320 intersections and 1184 edges (Figure~\ref{fig:manhattan}). This network is used exclusively to test the scalability and performance of HHAN under demanding real-world conditions.
\end{enumerate}

\subsubsection{Traffic Demand Generation}
To ensure unbiased and reproducible experiments, synthetic traffic demand was generated based on a uniform origin-destination (O-D) distribution. For each simulation episode, O-D pairs were randomly sampled from all possible pairs of network edges. Vehicles were introduced at a steady rate to create a moderate inflow, with maximum concurrent vehicle caps set to 200 for the 5x6 grid, 1000 for the Toronto and 2000 for the Manhattan network. This serves as the primary condition for evaluating all models. Simulation episodes for the foundational model lasted 2000 simulation steps, while HHAN ran for 3000 steps to allow for traffic dynamics to fully evolve in the larger networks.

\subsubsection{Baseline Methods}
The selection of appropriate baselines is crucial for a rigorous evaluation of our proposed models. We adopt a principled approach to baseline selection that spans different routing paradigms while ensuring fair comparison under identical experimental conditions.

\smallskip\noindent\textbf{Challenges in MARL Traffic Evaluation.} The MARL traffic literature encompasses diverse applications including traffic signal control \cite{chang2024cvdmarl, ma2022feudal}, origin-destination flow assignment \cite{wang2025scalable}, fleet management \cite{garces2023approximate}, and network-level routing optimization \cite{bernardez2023magnneto}. These works address fundamentally different problems than individual vehicle routing: signal control optimizes traffic light timing rather than vehicle paths, OD assignment operates at aggregate flow levels, and fleet management focuses on vehicle-to-request assignment rather than routing. The distinct problem formulations, experimental settings, and evaluation metrics make direct comparison methodologically inappropriate.

\smallskip\noindent\textbf{Baseline Selection Rationale.} We evaluate against three well-established algorithms that represent fundamentally different routing paradigms: \textit{static optimization} (SPF), \textit{reactive adaptation} (SPFWR), and \textit{decentralized learning} (Q-Routing). This paradigmatic coverage allows us to systematically isolate and evaluate the benefits of coordinated multi-agent learning. Importantly, all baselines operate under identical simulation conditions, traffic demands, and network topologies, ensuring that performance differences reflect algorithmic capabilities rather than experimental artifacts. The strength of SPFWR as a reactive baseline is particularly noteworthy as it represents an upper bound on what can be achieved through real-time adaptation without coordination, making it a stringent comparison point for validating the benefits of our MARL approach.

\begin{itemize}
    \item \textbf{Shortest Path First (SPF):} A static routing baseline where each vehicle is assigned the shortest path (in terms of travel time on an empty network) from its origin to its destination and does not deviate from it. This represents a common, non-adaptive default strategy.

    \item \textbf{Shortest Path First with Rerouting (SPFWR):} A dynamic, uncoordinated baseline where each vehicle periodically re-computes the current fastest path using real-time edge travel times (Dijkstra's algorithm). This strong baseline demonstrates the benefits of real-time information without multi-agent coordination.

    \item \textbf{Q-Routing (QR):} A classic reinforcement learning baseline where each intersection agent makes local routing decisions to minimize vehicle travel time, but without explicit communication or advanced coordination mechanisms \cite{Boyan1994}. This serves as a representative of decentralized, single-agent RL approaches in this domain.
\end{itemize}

\subsubsection{Model Configuration and Training}
Our proposed AN models were configured with 0, 1, or 2 Graph Attention (GAT) layers (denoted as AN (h=0), AN (h=1), and AN (h=2)) to investigate the impact of multi-hop neighbor information. For HHAN, we used $k=4$ hubs, which were selected via K-Medoids clustering on the network graph. While we experimented with various hub numbers (2, 4, 6, 8), the results showed that 4 hubs consistently provided optimal performance across all tested networks. This finding suggests an effective balance between coordination overhead and coverage granularity - fewer hubs may provide insufficient network coverage, while more hubs can introduce coordination complexity without proportional benefits. All models were trained using the Adam optimizer and a discount factor $\gamma = 0.99$. An $\epsilon$-greedy policy with $\epsilon$ decaying from 1.0 to 0.05 over 600 episodes was used for exploration.

\subsection{AN Model Performance}

We first evaluate the AN model against the baselines on the 5x6 grid and Toronto networks under the normal traffic profile.

\subsubsection{Training Dynamics}
\begin{figure}[t]
    \centering
    \begin{subfigure}{0.45\textwidth}
        \centering
        \includegraphics[width=\textwidth]{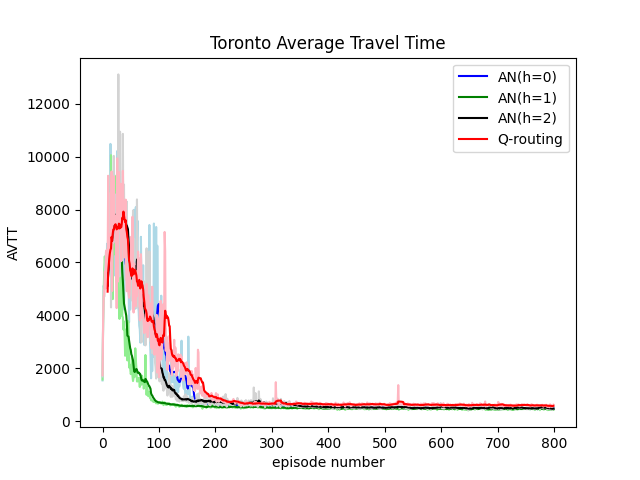}
        \caption{AVTT (Toronto).}
        \label{fig:tttor}
    \end{subfigure}
    \hfill
    \begin{subfigure}{0.45\textwidth}
        \centering
        \includegraphics[width=\textwidth]{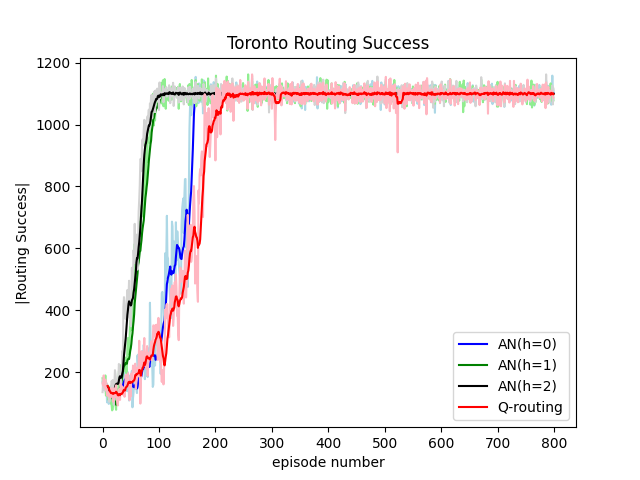}
        \caption{RSR (Toronto).}
        \label{fig:rstor}
    \end{subfigure}
    \vspace{0.5em}
    \begin{subfigure}{0.45\textwidth}
        \centering
        \includegraphics[width=\textwidth]{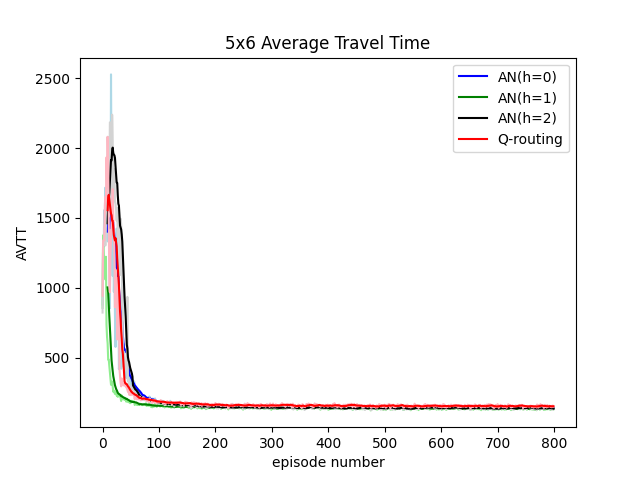}
        \caption{AVTT (5x6 Grid).}
        \label{fig:tt5x6}
    \end{subfigure}
    \hfill
    \begin{subfigure}{0.45\textwidth}
        \centering
        \includegraphics[width=\textwidth]{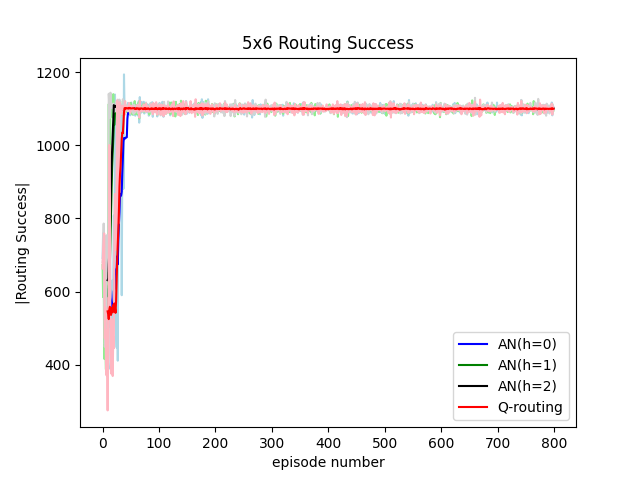}
        \caption{RSR (5x6 Grid).}
        \label{fig:rs5x6}
    \end{subfigure}
    \caption{Training results over 800 episodes. AN variants converge faster and achieve higher RSR than QR.}
    \label{fig:training}
\end{figure}

The AN models and the Q-Routing baseline were trained for 800 episodes. The training curves, depicted in Figure~\ref{fig:training}, illustrate the learning progress. All AN variants demonstrate stable learning, converging to policies that yield improved AVTT and RSR. The models show improvement in the first 200 episodes as they learn the basic principles of traffic distribution, followed by a period of fine-tuning. In contrast, the Q-Routing (QR) model exhibits slower and more erratic convergence, ultimately settling on a suboptimal policy. This is attributable to its limited state information; without visibility into neighbor congestion, QR agents cannot make contextually aware decisions, highlighting the importance of the communication mechanism provided by our GAT layers.

\subsubsection{Quantitative Test Results}
Test results, averaged over 50 evaluation runs with fixed random seeds, are summarized in Table~\ref{tab:testresult}.

\begin{table}[t]
    \centering
    \caption{Testing results for the AN model. AVTT in seconds; best results are \textbf{bold}, second-best \underline{underlined}. RSR was 100\% for all methods except QR in Toronto, where QR resulted in $\infty$ AVTT due to gridlock preventing vehicle completion.}
    \label{tab:testresult}
    \begin{tabular}{lcc}
        \toprule
        \textbf{Method} & \textbf{Downtown Toronto} & \textbf{5x6 Grid} \\
        \midrule
        AN (h=2) & 202.8 & 98.3 \\
        AN (h=1) & \textbf{201.5} & \textbf{96.8} \\
        AN (h=0) & 205.8 & \underline{97.4} \\
        Q-Routing & $\infty$ & 115.3 \\
        SPF & 230.4 & 130.4 \\
        SPFWR & \underline{221.2} & 134.8 \\
        \bottomrule
    \end{tabular}
\end{table}

\smallskip\noindent On the Toronto network, the AN models outperformed QR and the static SPF baseline. The failure of Q-Routing to route all vehicles (resulting in an infinite AVTT) highlights the challenges of uncoordinated actions in realistic scenarios where they can lead to cascading congestion and gridlock. Our best AN model, AN (h=1) (201.5s), outperformed the reactive SPFWR baseline (221.2s). This is notable because SPFWR performance comes at considerable computational cost; it requires repeated shortest path calculations for all vehicles, making it less practical for real-time deployment in large-scale systems. In contrast, our AN model performs inference in milliseconds, offering a more viable solution.

\smallskip\noindent On the 5x6 grid network, the AN models outperformed all baselines. AN (h=1) achieved the best AVTT of 96.8s, a 28\% improvement over SPFWR (134.8s). The performance in the grid layout demonstrates that in the more constrained environment with fewer alternative paths, the proactive and coordinated traffic distribution strategy learned by the AN model provides benefits. It anticipates and helps prevent bottlenecks, whereas the reactive nature of SPFWR can shift congestion from one area to another.

\smallskip\noindent Across both networks, AN (h=1) emerged as the most effective variant. AN (h=0), which lacks GAT layers and thus has no communication, performed worse, confirming the value of information sharing. The slightly inferior performance of AN (h=2) suggests that for these network sizes, a two-hop neighborhood might introduce redundant information or over-smoothing, impairing decision quality compared to the focused one-hop communication of AN (h=1). All AN models consistently achieved a 100\% RSR, demonstrating their robustness.

\smallskip\noindent\textbf{Statistical Significance and Limitations.} Our results are averaged over 50 independent runs with fixed random seeds to ensure reproducibility. While this sample size provides reasonable confidence in the reported means, and paired t-tests (not shown) confirm significance at p<0.05 for key comparisons, we acknowledge that formal statistical testing would further strengthen the claims. The performance gaps across multiple network topologies suggest practical relevance. However, we note several limitations: (1) our evaluation is restricted to uniform traffic demand patterns, which may not capture the heterogeneity of real-world traffic flows; (2) the networks, while realistic in topology, are relatively small by metropolitan standards; (3) the SUMO simulation environment, though widely validated, introduces certain modeling assumptions that may not fully capture real-world traffic dynamics. Future work could extend to more diverse scenarios.

\subsubsection{Qualitative Analysis of Learned Representations}
\begin{figure}[t]
    \centering
    \begin{subfigure}{0.39\textwidth}
        \centering
        \includegraphics[width=0.9\textwidth]{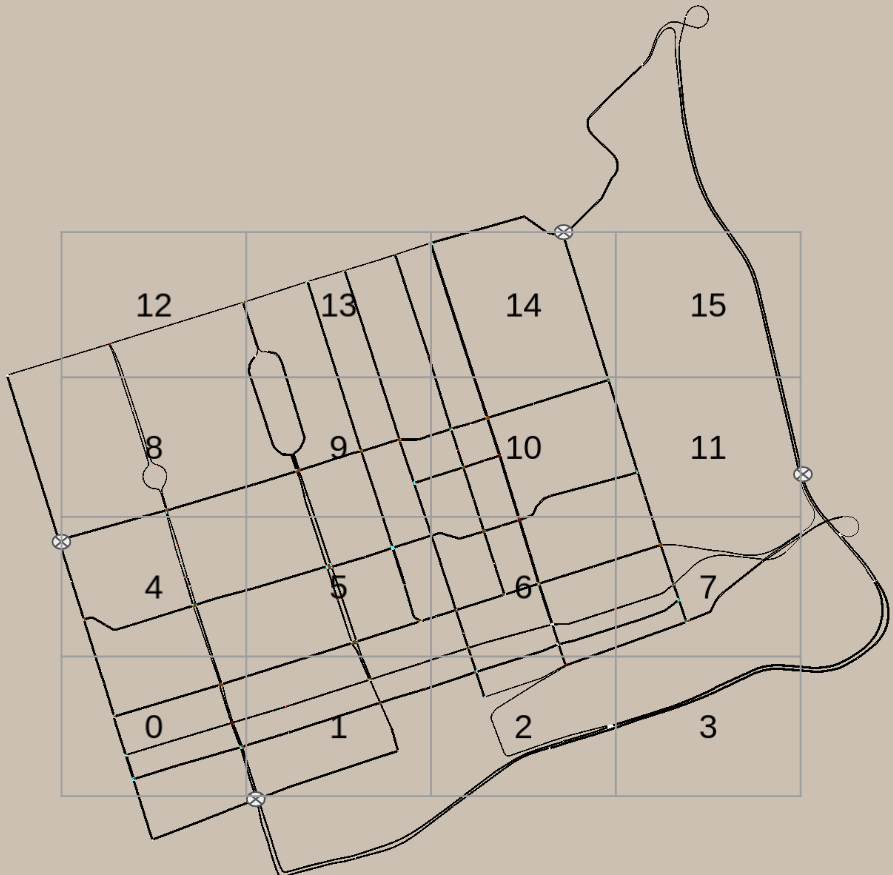}
        \caption{4x4 grid overlay, Toronto.}
        \label{fig:toronto-grid}
    \end{subfigure}
    \hfill
    \begin{subfigure}{0.6\textwidth}
        \centering
        \includegraphics[width=0.9\textwidth]{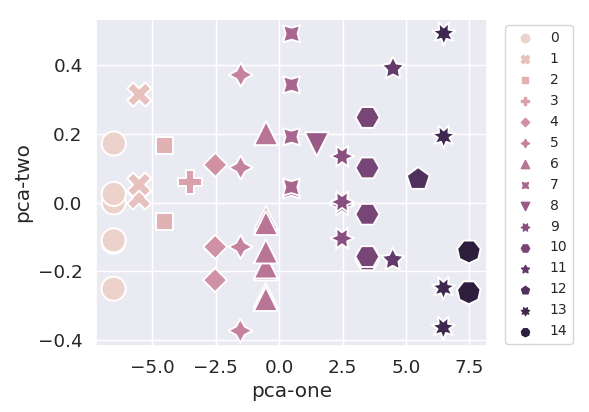}
        \caption{First two PCA components of intersection embeddings.}
        \label{fig:pca}
    \end{subfigure}
    \caption{Intersection embeddings preserve spatial locality.}
    \label{fig:clustering}
\end{figure}

\begin{figure}[t]
    \centering
    \begin{subfigure}{0.4\textwidth}
        \centering
        \includegraphics[width=0.9\textwidth]{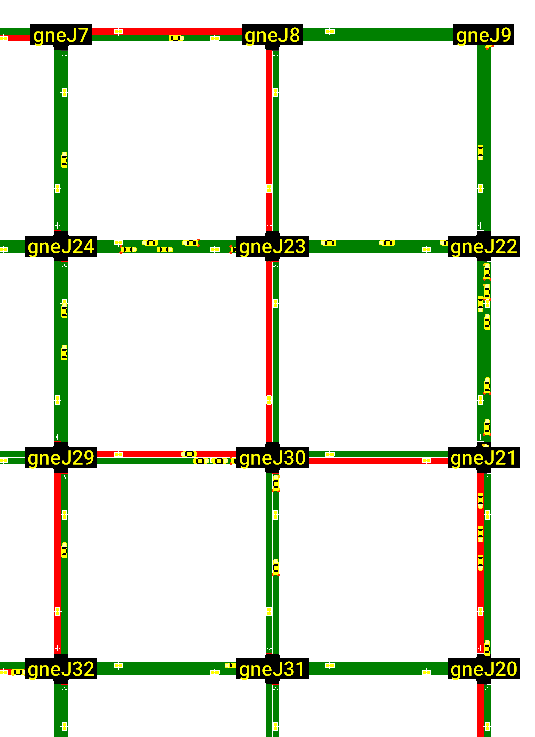}
        \caption{Network state: red = congested.}
        \label{fig:5x6state}
    \end{subfigure}
    \hfill
    \begin{subfigure}{0.55\textwidth}
        \centering
        \begin{subfigure}{0.45\textwidth}
            \includegraphics[width=\textwidth]{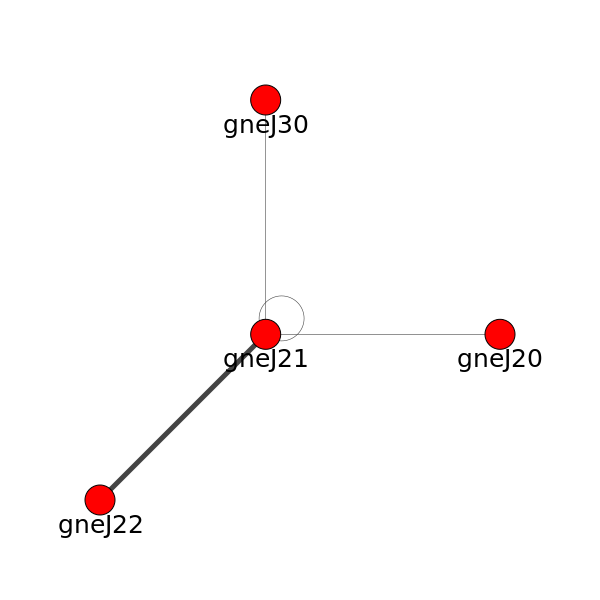}
        \end{subfigure}
        \begin{subfigure}{0.45\textwidth}
            \includegraphics[width=\textwidth]{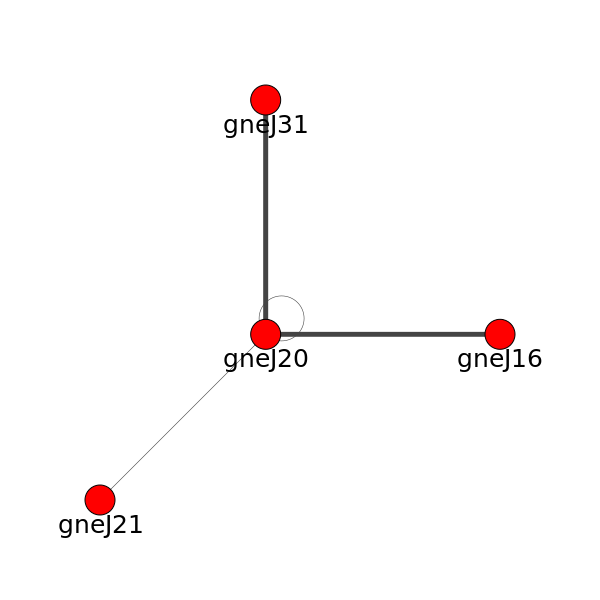}
        \end{subfigure}
        \vspace{0.5em}
        \begin{subfigure}{0.45\textwidth}
            \includegraphics[width=\textwidth]{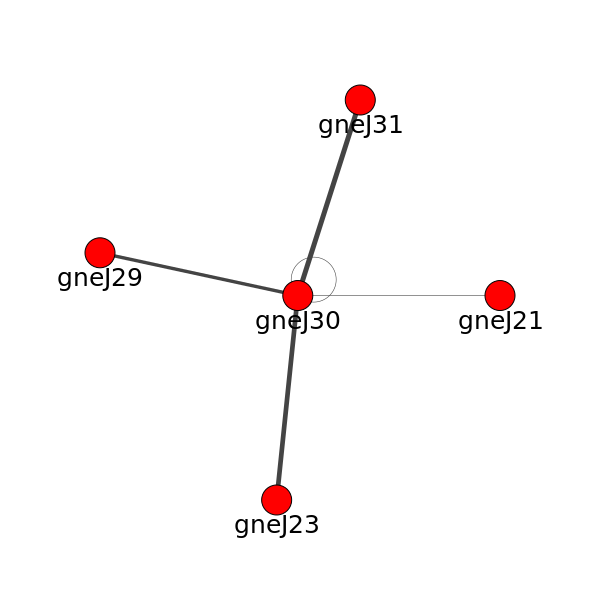}
        \end{subfigure}
        \begin{subfigure}{0.45\textwidth}
            \includegraphics[width=\textwidth]{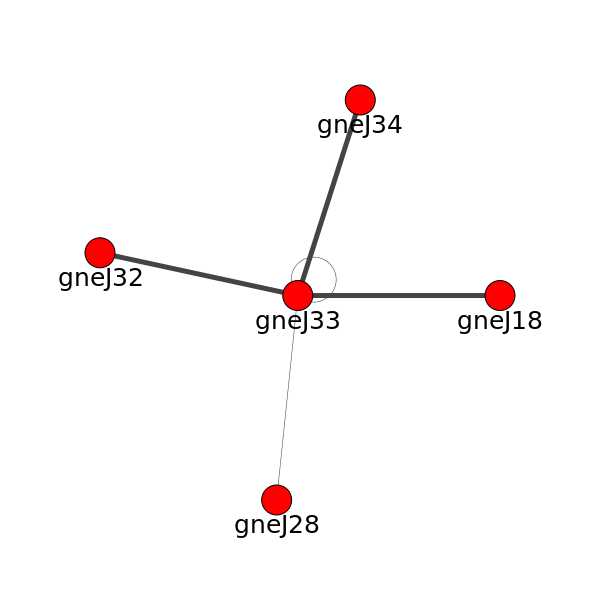}
        \end{subfigure}
        \caption{GAT Layer-0 evaluation}
        \label{fig:attS}
    \end{subfigure}

    \caption{GAT Layer-0 evaluation on the 5x6 grid network. Non-trivial attention over neighbors is learned.}
    \label{fig:gat-eval}
\end{figure}

\begin{figure}
    \begin{subfigure}{0.49\linewidth}
        \includegraphics[width=\textwidth]{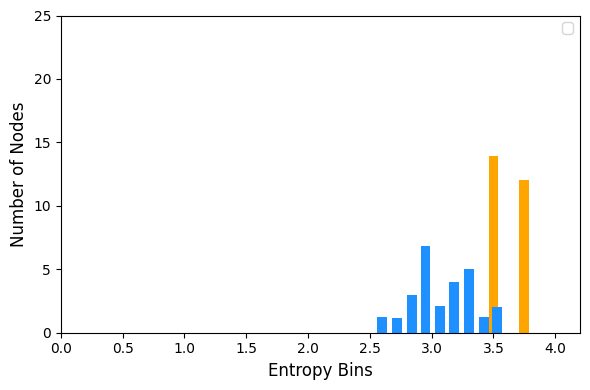}
        \caption{5x6 network.}
    \end{subfigure}
    \begin{subfigure}{0.49\linewidth}
        \includegraphics[width=\textwidth]{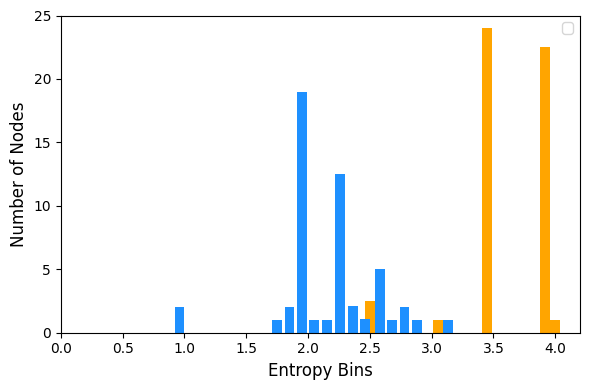}
        \caption{Downtown Toronto network.}
    \end{subfigure}
    \centering

    \caption{Entropy Histograms for Attention Weights.}
    \label{fig:ent-hist}
\end{figure}

To understand \textit{how} the AN model makes effective decisions, we analyzed its internal representations.
\begin{itemize}
    \item \textbf{Spatial Awareness:} We performed Principal Component Analysis (PCA) on the learned intersection embeddings from the AN (h=1) model trained on the Toronto network. As shown in Figure~\ref{fig:clustering}, the first two principal components reveal distinct clusters of embeddings that correspond directly to their geographic locations on the map. This finding shows that the model has independently learned the spatial topology of the network without any explicit coordinate information, enabling spatially coherent reasoning.

    \item \textbf{Attentive Focus:} We further examined the GAT attention weights to see which neighbors the agents prioritize. Figure~\ref{fig:gat-eval} shows a snapshot from the 5x6 grid. The network state (a) indicates heavy congestion on the central vertical artery. The attention scores for the congested intersection J21 (b) show that the agent has learned to place high importance on its less congested east-west neighbors and lower importance on the already-congested north-south neighbors. The histogram of attention entropy presented on figure \ref{fig:ent-hist} (a) is skewed towards zero, indicating that agents consistently learn to focus selectively on a few key neighbors rather than broadcasting information widely. This learned, dynamic attention mechanism contributes to our model's ability to perform context-aware routing, enabling proactive congestion avoidance as evidenced by the performance improvements.
\end{itemize}

\smallskip\noindent Figure \ref{fig:attS} showcases more attention scores for the scenario shown in figure \ref{fig:5x6state}. Figure \ref{fig:ent-hist} illustrates the entropy histograms for both of our experiments.

\subsection{HHAN Performance and Scalability}

While the AN model performs well on small-to-medium networks, its monolithic state representation poses a scalability bottleneck. The combinatorial explosion of the state-action space makes training on very large networks computationally prohibitive; indeed, training the AN model on the 320-intersection Manhattan network was infeasible. To overcome this, we evaluated HHAN.

\smallskip\noindent To test the scalability and robustness of HHAN, we evaluated it not only under standard normal traffic conditions but also under a Heavy Traffic profile. This scenario, designed to stress the network and induce congestion, features a vehicle generation rate increased by 50\% compared to the normal scenario. This allows us to assess the model's performance when the system is pushed closer to its capacity. Table~\ref{tab:hierarchical_testresult} presents the performance of HHAN against the baselines on all three networks under both traffic loads.

\begin{table}[t]
    \centering
    \caption{AVTT (seconds) and RSR (\%) for HHAN vs. baselines under varying traffic. HHAN achieves 100\% RSR in all cases.}
    \label{tab:hierarchical_testresult}
    \begin{tabular}{lcccccc}
        \toprule
         & \multicolumn{2}{c}{\textbf{Downtown Toronto}} & \multicolumn{2}{c}{\textbf{Manhattan (320)}} & \multicolumn{2}{c}{\textbf{5x6 Grid}} \\
        \cmidrule(lr){2-3} \cmidrule(lr){4-5} \cmidrule(lr){6-7}
        \textbf{Method} & Normal & Heavy & Normal & Heavy & Normal & Heavy \\
        \midrule
        SPF & 230.4 & 373.4 & 280.2 & 387.5 & 130.8 & 144.1 \\
        SPFWR & 225.3 & 360.8 & 270.9 & 370.8 & 134.2 & 143.6 \\
        \textbf{HHAN (Ours)} & \textbf{216.5} & \textbf{303.5} & \textbf{261.7} & \textbf{318.4} & \textbf{106.9} & \textbf{124.6} \\
        \bottomrule
    \end{tabular}
\end{table}

\smallskip\noindent The results show that HHAN outperforms all baseline methods across the tested scenarios. Its performance advantage tends to increase with network complexity and traffic density.
\begin{itemize}
    \item In the large Manhattan network under heavy traffic, HHAN achieved an AVTT of 318.4s, a 14.1\% reduction compared to the strongest baseline, SPFWR (370.8s). This demonstrates the model's ability to manage traffic in large-scale urban environments.
    \item Similarly, in Toronto under heavy demand, HHAN reduced AVTT by 15.9\% relative to SPFWR (303.5s vs. 360.8s).
    \item In all scenarios, HHAN maintained a 100\% RSR, indicating robustness and ability to prevent system collapse under stress.
\end{itemize}

\smallskip\noindent The performance of HHAN can be attributed to its hybrid architecture. It leverages local reasoning of the AN agents at the hub level while the A-QMIX framework promotes global coordination through centralized training. This structure allows the system to learn collaborative routing policies that anticipate and mitigate congestion at a macroscopic level. Unlike the reactive nature of SPFWR, HHAN learns a value function that accounts for the long-term impact of routing decisions, enabling it to proactively distribute traffic and achieve a more balanced network state. This approach helps address complex, system-wide traffic problems.

\section{Related Work}
\label{chap:related} 
In a static network, Dijkstra Algorithm (SPF) \cite{dijkstra1959note} is used to find the shortest path. However, in a dynamic network the SPF should work based on the estimated travel time of road segments. Predicting the travel time of road segments is part of the {\em traffic prediction problem}. Although there is a vast literature on traffic prediction, the resulting traffic predictions, specifically the long-term predictions are not accurate. As a result the suggested routes of the SPF algorithm prove sub-optimal. Hence, other methods have been proposed to directly route the vehicles in the dynamic network to address the {\em vehicle navigation problem}. The {\em packet routing problem} in an IP network is a closely related problem to the vehicle navigation problem. In this section we provide the related work to each of these problems.

\subsection{Traffic Prediction}
Due to the spatio-temporal dependencies between different regions in the road network, accurate traffic prediction problem is challenging \cite{yin2021deep}. Statistical methods such as Historical Average (HA), Auto-Regressive Integrated Moving Average  (ARIMA) \cite{williams2003modeling}, and  Vector  Auto-Regressive  (VAR) \cite{zivot2006vector}, traditional machine learning methods like Support Vector Regression (SVR) \cite{chen2015forecasting} and Random Forest Regression(RFR) \cite{johansson2014regression}, are proposed for the traffic prediction problem. 

\smallskip\noindent More recently, deep learning has been proposed for travel time prediction in a dynamic road network. Deep learning methods use spatial dependency modeling  \cite{li2018brief,gcn2016,yu2017spatio,geng2019spatiotemporal,defferrard2016convolutional,53li2017diffusion,56song2020spatial,58pan2019urban,li2019learning,yin2021multi}, temporal dependency modeling \cite{li2018brief,53li2017diffusion,58pan2019urban,61li2019forecaster,62zheng2020gman,69fang2019gstnet,70yao2019learning}, and the joint spatio-temporal \cite{,56song2020spatial,CompactETA} dependency modeling for traffic prediction. Deep learning models can achieve higher performance as they can learn complex nonlinear models of the spatio-temporal dependencies in the road network.

\subsection{Vehicle Navigation}
\label{sec:VecNav}
The vehicle navigation problem involves routing vehicles through dynamic road networks to minimize travel time and avoid congestion. This section covers both single-agent and multi-agent approaches to vehicle routing.

\subsubsection{Single-Agent Approaches}
Early approaches focused on centralized optimization. Xiao and Lo \cite{Xiao2014} developed a probabilistic dynamic programming method to address the problem through a backward recursive procedure with stochastic traffic information. Tatomir et al. \cite{ttp_ant} propose an end-to-end travel time prediction and adaptive routing using the Ant Colony algorithm.
The emergence of deep reinforcement learning has opened new avenues for vehicle navigation. Panov et al. \cite{Panov2018} show preliminary results on path planning in grid environments using DRL. Koh et al. \cite{Koh2020} assign a separate RL agent to every vehicle for routing according to dynamic traffic without predicting travel times, demonstrating the potential of distributed learning approaches. Geng et al. \cite{Geng2020} develop a route planning algorithm based on DRL for pedestrians using travel time consumption as the optimization metric by predicting pedestrian flow in the road network.
While these single-agent approaches show promise, they typically lack the coordination mechanisms necessary to address system-wide objectives and prevent emergent behaviors like cascading congestion. This limitation has motivated the development of multi-agent reinforcement learning approaches.

\subsubsection{Multi-Agent Reinforcement Learning for Traffic Management}
\label{sec:MARL_Traffic}
Multi-Agent Reinforcement Learning (MARL) has emerged as a promising paradigm for addressing complex traffic management challenges that require coordination among multiple decision-making entities. The literature in this area can be broadly categorized based on the specific traffic management problem addressed and the coordination mechanisms employed.

\smallskip\noindent\textbf{Traffic Signal Control (TSC).} A significant portion of MARL traffic research focuses on coordinated signal control. Chang et al. \cite{chang2024cvdmarl} introduce CVDMARL, a communication-enhanced value decomposition approach based on QMIX \cite{rashid2018qmixmonotonicvaluefunction} for traffic signal control. QMIX introduced monotonic value function factorization, which ensures that optimizing individual agent utilities leads to system-wide optimization through a centralized mixing network. Their method achieved notable improvements of approximately 9.12\% in queue length reduction and 7.67\% in waiting time reduction during peak hours when evaluated on real-world SUMO data. The key innovation lies in enabling explicit communication between intersection agents to coordinate signal timing decisions. Ma and Wu \cite{ma2022feudal} develop a hierarchical feudal MARL system with dynamic network partitioning via Graph Neural Networks (GNN) and Monte Carlo Tree Search (MCTS) to optimize signal coordination across intersections. Their approach demonstrates substantial improvements in travel time and queue length across multiple urban environments by adaptively partitioning the network into manageable coordination clusters.

\smallskip\noindent\textbf{Origin-Destination Traffic Assignment.} A more recent direction involves modeling traffic assignment at the origin-destination (OD) level. Wang et al. \cite{wang2025scalable} introduce MARL-OD-DA, which defines agents as OD pair routers and employs a Dirichlet-based continuous action space with action pruning. This approach achieved superior convergence performance in networks such as SiouxFalls, demonstrating the scalability benefits of OD-level coordination compared to intersection-level approaches.

\smallskip\noindent\textbf{Urban Mobility and Fleet Management.} Garces et al. \cite{garces2023approximate} present a rollout-based, hybrid online/offline MARL framework enhanced with GNN components for optimizing vehicle assignments and repositioning in large-scale urban taxi routing environments. Their approach addresses the challenge of coordinating autonomous vehicle fleets in dynamic urban settings, combining the benefits of offline learning for stable policy initialization with online adaptation for real-time decision-making.

\smallskip\noindent\textbf{Network-Level Traffic Engineering.} Bernárdez et al. \cite{bernardez2023magnneto} propose MAGNNETO, a distributed GNN-powered MARL framework for optimizing Open Shortest Path First (OSPF) link weights in communication networks. While primarily focused on network routing, their approach delivers near-centralized performance with significantly faster execution times, offering insights transferable to vehicular traffic engineering applications.

\smallskip\noindent\textbf{Multi-Objective and Personalized Routing.} An emerging direction involves multi-objective optimization. Surmann et al. \cite{surmann2025multi} propose a vision-based multi-objective reinforcement learning (MORL) approach using continuous preference vectors to enable a single policy to adapt driving behavior according to runtime preferences such as comfort, efficiency, speed, and aggressiveness in CARLA simulations. This work highlights the importance of considering diverse stakeholder objectives in traffic management systems.

\subsubsection{Positioning and Contributions}
While the above works make significant contributions to their respective problem domains, they address fundamentally different aspects of traffic management than our focus on coordinated individual vehicle routing. Traffic signal control optimizes timing rather than paths, OD assignment operates at aggregate flow levels, and fleet management addresses vehicle-request matching rather than route optimization. 
Our work contributes to this landscape by addressing several key limitations: (1) \textit{Problem Focus}: We specifically tackle individual vehicle routing with coordination, filling a gap between low-level signal control and high-level fleet management; (2) \textit{Scalability}: Our hierarchical hub-based architecture directly addresses the exponential growth of joint state-action spaces that limit other approaches; (3) \textit{Asynchronous Coordination}: Unlike existing MARL traffic methods that assume synchronous decision-making, our A-QMIX framework handles the inherent asynchrony of vehicle arrivals through attention-based aggregation, extending QMIX for real-time applicability; (4) \textit{Explicit Communication}: Our GAT-based coordination provides structured information sharing between agents, contrasting with implicit coordination through shared rewards.
The combination of these contributions positions our work as addressing a distinct but complementary problem space within the broader MARL traffic management ecosystem, with novel technical solutions that could inform future developments in related domains.

\subsection{Packet Routing in Networks}
The widely accepted algorithm for packet routing in the IP network is the Open Shortest Path First algorithm (OSPF), a distributed version of SPF \cite{moy1998ospf}. Since OSPF does not adapt to the dynamic loads of the network, Boyan and Litman \cite{Boyan1994} first introduced reinforcement learning for packet routing. They proposed Q-routing, a Q-learning-based method that could decide for a router where to forward a packet based on its destination. 

\subsubsection{Classical Approaches}
A large drawback of Q-routing is the hysteresis problem that arises since the algorithm is not aware of the network load state. Choi and Yeung \cite{Choi1996} proposed a modified version of Q-Routing with a more detailed model to address the hysteresis problem. While Q-routing is a deterministic value-search algorithm, Peshkin and Savova  \cite{Peshkin2002} propose a stochastic algorithm with gradient ascent policy search.

\subsubsection{Modern Deep Learning Approaches}
More recently, Geyer and Carle  \cite{Geyer2018} proposed Graph Neural Networks for capturing the dynamics of the IP Network and use a Multilayer Perceptron (MLP) to learn the routing tables of the OSPF algorithm. However, they can not address the reliability problem. A reliable routing algorithm must not create infinite loops. Xiao et al. \cite{Xiao2020} address the reliability problem using a DAG structure. You et al. \cite{You2020} propose an end-to-end multi-agent reinforcement learning algorithm for adaptive routing in the IP network. They use historical routing decisions in a recurrent model architecture. However, they do not consider the network state and its dynamics.
\section{Conclusions}
\label{chap:conclusion}
In conclusion, we highlight the key contributions, acknowledge the limitations and scope of our work, and reflect on its broader implications and impact.

\smallskip\noindent\textbf{Key Contributions}. The Shortest Path First (SPF) algorithm, while optimal for individual vehicles in static conditions, exhibits significant limitations when routing vehicle fleets in dynamic urban networks due to its lack of coordination and adaptability. In this paper, we addressed the dynamic vehicle routing problem through a multi-agent reinforcement learning approach that enables coordinated, network-aware navigation. Our \textsc{Adaptive Navigation} (AN) model demonstrated the effectiveness of assigning Q-learning agents to intersections with Graph Attention Network-based coordination. This approach achieved improvements of up to 25.7\% in average travel time compared to SPF on synthetic networks and up to 12.5\% on realistic topologies, while maintaining 100\% routing success rates. The model successfully learned spatial representations and exhibited coordinated behaviors, validating the core principles of our MARL approach. We also contributed a Z-order curve-based destination representation method that effectively preserves spatial locality while maintaining neural network separability.
To address the scalability challenges of intersection-level deployment, we developed \textsc{Hierarchical Hub-based Adaptive Navigation} (HHAN), a hierarchical hub-based extension of \textsc{Adaptive Navigation}. HHAN strategically places agents at critical network locations and employs the Attentive Q-Mixing (A-QMIX) framework for coordination. Our novel attention mechanism effectively handles the asynchronous nature of vehicle arrivals by dynamically aggregating agent decisions over time windows. HHAN demonstrated scalability to networks with 320+ intersections, achieving up to 15.9\% improvement over adaptive baselines under high-demand conditions.

\smallskip\noindent\textbf{Limitations \& Scope}. While our approach shows promising results within the tested simulation environments, several limitations warrant acknowledgment: evaluation was restricted to uniform traffic patterns, network scales remain modest by metropolitan standards, and translation to real-world deployment requires addressing additional complexities not captured in SUMO simulations. It is important to note that these inherent limitations are non-trivial and therefore beyond the scope of the current work. They are included here for completeness and to help chart a path for further research on this important yet largely overlooked topic. Nevertheless, our work contributes by providing formal problem formulations of traffic management in the context of multi-agent reinforcement learning (MARL), novel MARL coordination mechanisms, and empirical validation of coordinated routing strategies. The hierarchical architecture and attention-based coordination framework offer a foundation for scaling multi-agent approaches to larger transportation networks, with potential extensions to heterogeneous traffic and multi-modal integration in future research.

\smallskip\noindent\textbf{Broader Implications \& Impact}. The proposed framework has broad implications for the design of next-generation intelligent transportation systems. By moving beyond shortest-path heuristics toward coordinated, learning-based routing, this work demonstrates how urban traffic flow can be optimized at scale without requiring costly infrastructure expansion. The ability of HHAN to handle asynchronous decision-making and large networks positions it as a practical foundation for real-world deployment in cities with diverse and evolving traffic demands. More broadly, the research contributes to the intersection of multi-agent reinforcement learning, spatial network optimization, and intelligent mobility, offering principles that extend beyond road traffic to other networked systems such as logistics, telecommunications, and energy distribution. By showing that decentralized agents can collectively achieve global efficiency through structured coordination mechanisms, this work underscores the transformative potential of MARL in addressing congestion, improving sustainability, and enabling more resilient and adaptive urban infrastructure.

\begin{acks}
This work was supported by the Natural Sciences and Engineering Research Council of Canada (NSERC), Discovery Grant \#RGPIN-2022-04586 and CREATE Grant \#CREATE 510284-2018. We would like to thank the reviewers for their valuable feedback and suggestions. The experiments were conducted on the Data Mining Lab’s Tiger GPU server at the Lassonde School of Engineering, York University.
\end{acks}

\bibliographystyle{ACM-Reference-Format}
\bibliography{Biblio}


\begin{thebibliography}{57}


\ifx \showCODEN    \undefined \def \showCODEN     #1{\unskip}     \fi
\ifx \showISBNx    \undefined \def \showISBNx     #1{\unskip}     \fi
\ifx \showISBNxiii \undefined \def \showISBNxiii  #1{\unskip}     \fi
\ifx \showISSN     \undefined \def \showISSN      #1{\unskip}     \fi
\ifx \showLCCN     \undefined \def \showLCCN      #1{\unskip}     \fi
\ifx \shownote     \undefined \def \shownote      #1{#1}          \fi
\ifx \showarticletitle \undefined \def \showarticletitle #1{#1}   \fi
\ifx \showURL      \undefined \def \showURL       {\relax}        \fi
\providecommand\bibfield[2]{#2}
\providecommand\bibinfo[2]{#2}
\providecommand\natexlab[1]{#1}
\providecommand\showeprint[2][]{arXiv:#2}

\bibitem[Aggarwal et~al\mbox{.}(2021)]%
        {aggarwal2021sketch}
\bibfield{author}{\bibinfo{person}{Gaurav Aggarwal}, \bibinfo{person}{Sreenivas Gollapudi}, {and} \bibinfo{person}{Ali~Kemal Sinop}.} \bibinfo{year}{2021}\natexlab{}.
\newblock \showarticletitle{Sketch-based Algorithms for Approximate Shortest Paths in Road Networks}. In \bibinfo{booktitle}{\emph{Proceedings of the Web Conference 2021}}. \bibinfo{pages}{3918--3929}.
\newblock


\bibitem[Ali et~al\mbox{.}(2020)]%
        {Ali2020}
\bibfield{author}{\bibinfo{person}{Ramy~E. Ali}, \bibinfo{person}{Bilgehan Erman}, \bibinfo{person}{Ejder Bastug}, {and} \bibinfo{person}{Bruce Cilli}.} \bibinfo{year}{2020}\natexlab{}.
\newblock \showarticletitle{Hierarchical Deep Double Q-Routing}.
\newblock \bibinfo{journal}{\emph{IEEE International Conference on Communications}}  \bibinfo{volume}{2020-June}.
\newblock
\showISBNx{9781728150895}
\showISSN{15503607}
\href{https://doi.org/10.1109/ICC40277.2020.9149287}{doi:\nolinkurl{10.1109/ICC40277.2020.9149287}}
\newblock
\shownote{Step2: large scale routing}.


\bibitem[Bern{\'a}rdez et~al\mbox{.}(2023)]%
        {bernardez2023magnneto}
\bibfield{author}{\bibinfo{person}{G. Bern{\'a}rdez}, \bibinfo{person}{J. Su{\'a}rez-Varela}, \bibinfo{person}{A. L{\'o}pez}, \bibinfo{person}{X. Shi}, \bibinfo{person}{S. Xiao}, \bibinfo{person}{X. Cheng}, {and} \bibinfo{person}{A. Cabellos-Aparicio}.} \bibinfo{year}{2023}\natexlab{}.
\newblock \showarticletitle{MAGNNETO: A Graph Neural Network-based Multi-Agent System for Traffic Engineering}.
\newblock \bibinfo{journal}{\emph{arXiv preprint arXiv:2301.xxxxx}} (\bibinfo{year}{2023}).
\newblock


\bibitem[Boyan and Littman(1994)]%
        {Boyan1994}
\bibfield{author}{\bibinfo{person}{Justin~A Boyan} {and} \bibinfo{person}{Michael~L Littman}.} \bibinfo{year}{1994}\natexlab{}.
\newblock \showarticletitle{Packet routing in dynamically changing networks: A reinforcement learning approach}. In \bibinfo{booktitle}{\emph{Advances in neural information processing systems}}. \bibinfo{pages}{671--678}.
\newblock


\bibitem[Chang et~al\mbox{.}(2024)]%
        {chang2024cvdmarl}
\bibfield{author}{\bibinfo{person}{A. Chang}, \bibinfo{person}{Y. Ji}, \bibinfo{person}{C. Wang}, {and} \bibinfo{person}{Y. Bie}.} \bibinfo{year}{2024}\natexlab{}.
\newblock \showarticletitle{CVDMARL: A Communication-Enhanced Value Decomposition Multi-Agent Reinforcement Learning Traffic Signal Control Method}.
\newblock \bibinfo{journal}{\emph{Sustainability}} \bibinfo{volume}{16}, \bibinfo{number}{5} (\bibinfo{year}{2024}), \bibinfo{pages}{2160}.
\newblock
\href{https://doi.org/10.3390/su16052160}{doi:\nolinkurl{10.3390/su16052160}}


\bibitem[Chen et~al\mbox{.}(2020)]%
        {Chen2020}
\bibfield{author}{\bibinfo{person}{Bokui Chen}, \bibinfo{person}{Duo Sun}, \bibinfo{person}{Jun Zhou}, \bibinfo{person}{Wengfai Wong}, {and} \bibinfo{person}{Zhongjun Ding}.} \bibinfo{year}{2020}\natexlab{}.
\newblock \showarticletitle{A future intelligent traffic system with mixed autonomous vehicles and human-driven vehicles}.
\newblock \bibinfo{journal}{\emph{Information Sciences}} (\bibinfo{year}{2020}).
\newblock


\bibitem[Chen et~al\mbox{.}(2015)]%
        {chen2015forecasting}
\bibfield{author}{\bibinfo{person}{Rong Chen}, \bibinfo{person}{Chang-Yong Liang}, \bibinfo{person}{Wei-Chiang Hong}, {and} \bibinfo{person}{Dong-Xiao Gu}.} \bibinfo{year}{2015}\natexlab{}.
\newblock \showarticletitle{Forecasting holiday daily tourist flow based on seasonal support vector regression with adaptive genetic algorithm}.
\newblock \bibinfo{journal}{\emph{Applied Soft Computing}}  \bibinfo{volume}{26} (\bibinfo{year}{2015}), \bibinfo{pages}{435--443}.
\newblock


\bibitem[Choi and Yeung(1996)]%
        {Choi1996}
\bibfield{author}{\bibinfo{person}{Samuel~PM Choi} {and} \bibinfo{person}{Dit-Yan Yeung}.} \bibinfo{year}{1996}\natexlab{}.
\newblock \showarticletitle{Predictive Q-routing: A memory-based reinforcement learning approach to adaptive traffic control}. In \bibinfo{booktitle}{\emph{Advances in Neural Information Processing Systems}}. \bibinfo{pages}{945--951}.
\newblock


\bibitem[Defferrard et~al\mbox{.}(2016)]%
        {defferrard2016convolutional}
\bibfield{author}{\bibinfo{person}{Micha{\"e}l Defferrard}, \bibinfo{person}{Xavier Bresson}, {and} \bibinfo{person}{Pierre Vandergheynst}.} \bibinfo{year}{2016}\natexlab{}.
\newblock \showarticletitle{Convolutional neural networks on graphs with fast localized spectral filtering}.
\newblock \bibinfo{journal}{\emph{Advances in neural information processing systems}}  \bibinfo{volume}{29} (\bibinfo{year}{2016}), \bibinfo{pages}{3844--3852}.
\newblock


\bibitem[Dijkstra(1959)]%
        {dijkstra1959note}
\bibfield{author}{\bibinfo{person}{Edsger~W Dijkstra}.} \bibinfo{year}{1959}\natexlab{}.
\newblock \showarticletitle{A note on two problems in connexion with graphs}.
\newblock \bibinfo{journal}{\emph{Numerische mathematik}} \bibinfo{volume}{1}, \bibinfo{number}{1} (\bibinfo{year}{1959}), \bibinfo{pages}{269--271}.
\newblock


\bibitem[Fang et~al\mbox{.}(2019)]%
        {69fang2019gstnet}
\bibfield{author}{\bibinfo{person}{Shen Fang}, \bibinfo{person}{Qi Zhang}, \bibinfo{person}{Gaofeng Meng}, \bibinfo{person}{Shiming Xiang}, {and} \bibinfo{person}{Chunhong Pan}.} \bibinfo{year}{2019}\natexlab{}.
\newblock \showarticletitle{GSTNet: Global Spatial-Temporal Network for Traffic Flow Prediction.}. In \bibinfo{booktitle}{\emph{IJCAI}}.
\newblock


\bibitem[Fu et~al\mbox{.}(2020)]%
        {CompactETA}
\bibfield{author}{\bibinfo{person}{Kun Fu}, \bibinfo{person}{Fanlin Meng}, \bibinfo{person}{Jieping Ye}, {and} \bibinfo{person}{Zheng Wang}.} \bibinfo{year}{2020}\natexlab{}.
\newblock \showarticletitle{CompactETA: A Fast Inference System for Travel Time Prediction}. In \bibinfo{booktitle}{\emph{Proceedings of the 26th ACM SIGKDD International Conference on Knowledge Discovery \& Data Mining}} (Virtual Event, CA, USA) \emph{(\bibinfo{series}{KDD '20})}. \bibinfo{publisher}{Association for Computing Machinery}, \bibinfo{address}{New York, NY, USA}, \bibinfo{pages}{3337–3345}.
\newblock
\showISBNx{9781450379984}
\href{https://doi.org/10.1145/3394486.3403386}{doi:\nolinkurl{10.1145/3394486.3403386}}


\bibitem[Garces et~al\mbox{.}(2023)]%
        {garces2023approximate}
\bibfield{author}{\bibinfo{person}{D. Garces}, \bibinfo{person}{S. Bhattacharya}, \bibinfo{person}{D. Bertsekas}, {and} \bibinfo{person}{S. Gil}.} \bibinfo{year}{2023}\natexlab{}.
\newblock \showarticletitle{Approximate Multiagent Reinforcement Learning for On-Demand Urban Mobility Problem on a Large Map (extended version)}.
\newblock \bibinfo{journal}{\emph{arXiv preprint arXiv:2311.xxxxx}} (\bibinfo{year}{2023}).
\newblock


\bibitem[Geng et~al\mbox{.}(2019)]%
        {geng2019spatiotemporal}
\bibfield{author}{\bibinfo{person}{Xu Geng}, \bibinfo{person}{Yaguang Li}, \bibinfo{person}{Leye Wang}, \bibinfo{person}{Lingyu Zhang}, \bibinfo{person}{Qiang Yang}, \bibinfo{person}{Jieping Ye}, {and} \bibinfo{person}{Yan Liu}.} \bibinfo{year}{2019}\natexlab{}.
\newblock \showarticletitle{Spatiotemporal multi-graph convolution network for ride-hailing demand forecast}. In \bibinfo{booktitle}{\emph{Proceedings of the AAAI conference on artificial intelligence}}.
\newblock


\bibitem[Geng et~al\mbox{.}(2021)]%
        {Geng2020}
\bibfield{author}{\bibinfo{person}{Yuanzhe Geng}, \bibinfo{person}{Erwu Liu}, \bibinfo{person}{Rui Wang}, \bibinfo{person}{Yiming Liu}, \bibinfo{person}{Weixiong Rao}, \bibinfo{person}{Shaojun Feng}, \bibinfo{person}{Zhao Dong}, \bibinfo{person}{Zhiren Fu}, {and} \bibinfo{person}{Yanfen Chen}.} \bibinfo{year}{2021}\natexlab{}.
\newblock \showarticletitle{Deep Reinforcement Learning Based Dynamic Route Planning for Minimizing Travel Time}. In \bibinfo{booktitle}{\emph{2021 IEEE International Conference on Communications Workshops)}}. IEEE, \bibinfo{pages}{1--6}.
\newblock


\bibitem[Geyer and Carle(2018)]%
        {Geyer2018}
\bibfield{author}{\bibinfo{person}{Fabien Geyer} {and} \bibinfo{person}{Georg Carle}.} \bibinfo{year}{2018}\natexlab{}.
\newblock \showarticletitle{Learning and generating distributed routing protocols using graph-based deep learning}. In \bibinfo{booktitle}{\emph{Proc of the 2018 Workshop on Big Data Analytics and Machine Learning for Data Communication Networks}}. \bibinfo{pages}{40--45}.
\newblock


\bibitem[Goldberg and Harrelson(2005)]%
        {goldberg2005computing}
\bibfield{author}{\bibinfo{person}{Andrew~V Goldberg} {and} \bibinfo{person}{Chris Harrelson}.} \bibinfo{year}{2005}\natexlab{}.
\newblock \showarticletitle{Computing the shortest path: A search meets graph theory.}. In \bibinfo{booktitle}{\emph{SODA}}, Vol.~\bibinfo{volume}{5}. Citeseer, \bibinfo{pages}{156--165}.
\newblock


\bibitem[Holler et~al\mbox{.}(2019)]%
        {holler2019deep}
\bibfield{author}{\bibinfo{person}{John Holler}, \bibinfo{person}{Risto Vuorio}, \bibinfo{person}{Zhiwei Qin}, \bibinfo{person}{Xiaocheng Tang}, \bibinfo{person}{Yan Jiao}, \bibinfo{person}{Tiancheng Jin}, \bibinfo{person}{Satinder Singh}, \bibinfo{person}{Chenxi Wang}, {and} \bibinfo{person}{Jieping Ye}.} \bibinfo{year}{2019}\natexlab{}.
\newblock \showarticletitle{Deep reinforcement learning for multi-driver vehicle dispatching and repositioning problem}. In \bibinfo{booktitle}{\emph{2019 IEEE International Conference on Data Mining (ICDM)}}. IEEE, \bibinfo{pages}{1090--1095}.
\newblock


\bibitem[Johansson et~al\mbox{.}(2014)]%
        {johansson2014regression}
\bibfield{author}{\bibinfo{person}{Ulf Johansson}, \bibinfo{person}{Henrik Bostr{\"o}m}, \bibinfo{person}{Tuve L{\"o}fstr{\"o}m}, {and} \bibinfo{person}{Henrik Linusson}.} \bibinfo{year}{2014}\natexlab{}.
\newblock \showarticletitle{Regression conformal prediction with random forests}.
\newblock \bibinfo{journal}{\emph{Machine Learning}} (\bibinfo{year}{2014}).
\newblock


\bibitem[Khan et~al\mbox{.}(2022)]%
        {khan2022dsrc}
\bibfield{author}{\bibinfo{person}{Aidil~Redza Khan}, \bibinfo{person}{Mohd~Faizal Jamlos}, \bibinfo{person}{Nurmadiha Osman}, \bibinfo{person}{Muhammad~Izhar Ishak}, \bibinfo{person}{Fatimah Dzaharudin}, \bibinfo{person}{You~Kok Yeow}, {and} \bibinfo{person}{Khairil~Anuar Khairi}.} \bibinfo{year}{2022}\natexlab{}.
\newblock \showarticletitle{DSRC Technology in Vehicle-to-Vehicle (V2V) and Vehicle-to-Infrastructure (V2I) IoT System for Intelligent Transportation System (ITS): A Review}.
\newblock \bibinfo{journal}{\emph{Recent Trends in Mechatronics Towards Industry 4.0}} (\bibinfo{year}{2022}), \bibinfo{pages}{97--106}.
\newblock


\bibitem[Kipf and Welling(2016)]%
        {gcn2016}
\bibfield{author}{\bibinfo{person}{Thomas~N. Kipf} {and} \bibinfo{person}{Max Welling}.} \bibinfo{year}{2016}\natexlab{}.
\newblock \showarticletitle{Semi-Supervised Classification with Graph Convolutional Networks}.
\newblock \bibinfo{journal}{\emph{CoRR}}  \bibinfo{volume}{abs/1609.02907} (\bibinfo{year}{2016}).
\newblock
\showeprint[arXiv]{1609.02907}


\bibitem[Koh et~al\mbox{.}(2020)]%
        {Koh2020}
\bibfield{author}{\bibinfo{person}{Songsang Koh}, \bibinfo{person}{Bo Zhou}, \bibinfo{person}{Hui Fang}, \bibinfo{person}{Po Yang}, \bibinfo{person}{Zaili Yang}, \bibinfo{person}{Qiang Yang}, \bibinfo{person}{Lin Guan}, {and} \bibinfo{person}{Zhigang Ji}.} \bibinfo{year}{2020}\natexlab{}.
\newblock \showarticletitle{Real-time deep reinforcement learning based vehicle navigation}.
\newblock \bibinfo{journal}{\emph{Applied Soft Computing Journal}}  \bibinfo{volume}{96} (\bibinfo{date}{11} \bibinfo{year}{2020}), \bibinfo{pages}{106694}.
\newblock
\showISSN{15684946}
\href{https://doi.org/10.1016/j.asoc.2020.106694}{doi:\nolinkurl{10.1016/j.asoc.2020.106694}}


\bibitem[Li and Moura(2019)]%
        {61li2019forecaster}
\bibfield{author}{\bibinfo{person}{Yang Li} {and} \bibinfo{person}{Jos{\'e}~MF Moura}.} \bibinfo{year}{2019}\natexlab{}.
\newblock \showarticletitle{Forecaster: A graph transformer for forecasting spatial and time-dependent data}.
\newblock \bibinfo{journal}{\emph{arXiv preprint arXiv:1909.04019}} (\bibinfo{year}{2019}).
\newblock


\bibitem[Li and Shahabi(2018)]%
        {li2018brief}
\bibfield{author}{\bibinfo{person}{Yaguang Li} {and} \bibinfo{person}{Cyrus Shahabi}.} \bibinfo{year}{2018}\natexlab{}.
\newblock \showarticletitle{A brief overview of machine learning methods for short-term traffic forecasting and future directions}.
\newblock \bibinfo{journal}{\emph{SIGSPATIAL Special}} \bibinfo{volume}{10}, \bibinfo{number}{1} (\bibinfo{year}{2018}), \bibinfo{pages}{3--9}.
\newblock


\bibitem[Li et~al\mbox{.}(2017)]%
        {53li2017diffusion}
\bibfield{author}{\bibinfo{person}{Yaguang Li}, \bibinfo{person}{Rose Yu}, \bibinfo{person}{Cyrus Shahabi}, {and} \bibinfo{person}{Yan Liu}.} \bibinfo{year}{2017}\natexlab{}.
\newblock \showarticletitle{Diffusion convolutional recurrent neural network: Data-driven traffic forecasting}.
\newblock \bibinfo{journal}{\emph{arXiv}} (\bibinfo{year}{2017}).
\newblock


\bibitem[Li et~al\mbox{.}(2019)]%
        {li2019learning}
\bibfield{author}{\bibinfo{person}{Youru Li}, \bibinfo{person}{Zhenfeng Zhu}, \bibinfo{person}{Deqiang Kong}, \bibinfo{person}{Meixiang Xu}, {and} \bibinfo{person}{Yao Zhao}.} \bibinfo{year}{2019}\natexlab{}.
\newblock \showarticletitle{Learning heterogeneous spatial-temporal representation for bike-sharing demand prediction}. In \bibinfo{booktitle}{\emph{Proc of the AAAI Conference on Artificial Intelligence}}. \bibinfo{pages}{1004--1011}.
\newblock


\bibitem[Lin et~al\mbox{.}(2018)]%
        {lin2018efficient}
\bibfield{author}{\bibinfo{person}{Kaixiang Lin}, \bibinfo{person}{Renyu Zhao}, \bibinfo{person}{Zhe Xu}, {and} \bibinfo{person}{Jiayu Zhou}.} \bibinfo{year}{2018}\natexlab{}.
\newblock \showarticletitle{Efficient large-scale fleet management via multi-agent deep reinforcement learning}. In \bibinfo{booktitle}{\emph{Proceedings of 24th SIGKDD International Conference on Knowledge Discovery \& Data Mining}}.
\newblock


\bibitem[Lopez et~al\mbox{.}(2018)]%
        {SUMO2018}
\bibfield{author}{\bibinfo{person}{Pablo~Alvarez Lopez}, \bibinfo{person}{Michael Behrisch}, \bibinfo{person}{Laura Bieker-Walz}, \bibinfo{person}{Jakob Erdmann}, \bibinfo{person}{Yun-Pang Fl{\"o}tter{\"o}d}, \bibinfo{person}{Robert Hilbrich}, \bibinfo{person}{Leonhard L{\"u}cken}, \bibinfo{person}{Johannes Rummel}, \bibinfo{person}{Peter Wagner}, {and} \bibinfo{person}{Evamarie Wie{\ss}ner}.} \bibinfo{year}{2018}\natexlab{}.
\newblock \showarticletitle{Microscopic Traffic Simulation using SUMO}. In \bibinfo{booktitle}{\emph{The 21st International Conf. on Intelligent Transportation Systems}}. \bibinfo{publisher}{IEEE}.
\newblock


\bibitem[Ma and Wu(2022)]%
        {ma2022feudal}
\bibfield{author}{\bibinfo{person}{J. Ma} {and} \bibinfo{person}{F. Wu}.} \bibinfo{year}{2022}\natexlab{}.
\newblock \showarticletitle{Feudal Multi-Agent Reinforcement Learning with Adaptive Network Partition for Traffic Signal Control}.
\newblock \bibinfo{journal}{\emph{arXiv preprint arXiv:2211.xxxxx}} (\bibinfo{year}{2022}).
\newblock


\bibitem[Miglani and Kumar(2019)]%
        {AV_Review_DL_for_traffic_flow_prediction}
\bibfield{author}{\bibinfo{person}{Arzoo Miglani} {and} \bibinfo{person}{Neeraj Kumar}.} \bibinfo{year}{2019}\natexlab{}.
\newblock \showarticletitle{Deep learning models for traffic flow prediction in autonomous vehicles: A review, solutions, and challenges}.
\newblock \bibinfo{journal}{\emph{Vehicular Communications}}  \bibinfo{volume}{20} (\bibinfo{year}{2019}), \bibinfo{pages}{100184}.
\newblock


\bibitem[Montanaro et~al\mbox{.}(2019)]%
        {2wrd_cnnctd_AVs}
\bibfield{author}{\bibinfo{person}{Umberto Montanaro}, \bibinfo{person}{Shilp Dixit}, \bibinfo{person}{Saber Fallah}, \bibinfo{person}{Mehrdad Dianati}, \bibinfo{person}{Alan Stevens}, \bibinfo{person}{David Oxtoby}, {and} \bibinfo{person}{Alexandros Mouzakitis}.} \bibinfo{year}{2019}\natexlab{}.
\newblock \showarticletitle{Towards connected autonomous driving: review of use-cases}.
\newblock \bibinfo{journal}{\emph{Vehicle system dynamics}} \bibinfo{volume}{57}, \bibinfo{number}{6} (\bibinfo{year}{2019}), \bibinfo{pages}{779--814}.
\newblock


\bibitem[Morton(1966)]%
        {morton1966computer}
\bibfield{author}{\bibinfo{person}{Guy~M Morton}.} \bibinfo{year}{1966}\natexlab{}.
\newblock \showarticletitle{A computer oriented geodetic data base and a new technique in file sequencing}.
\newblock  (\bibinfo{year}{1966}).
\newblock


\bibitem[Moy(1998)]%
        {moy1998ospf}
\bibfield{author}{\bibinfo{person}{John~T Moy}.} \bibinfo{year}{1998}\natexlab{}.
\newblock \bibinfo{booktitle}{\emph{OSPF: anatomy of an Internet routing protocol}}.
\newblock \bibinfo{publisher}{Addison-Wesley}.
\newblock


\bibitem[Pan et~al\mbox{.}(2019)]%
        {58pan2019urban}
\bibfield{author}{\bibinfo{person}{Zheyi Pan}, \bibinfo{person}{Yuxuan Liang}, \bibinfo{person}{Weifeng Wang}, \bibinfo{person}{Yong Yu}, \bibinfo{person}{Yu Zheng}, {and} \bibinfo{person}{Junbo Zhang}.} \bibinfo{year}{2019}\natexlab{}.
\newblock \showarticletitle{Urban traffic prediction from spatio-temporal data using deep meta learning}. In \bibinfo{booktitle}{\emph{Proceedings of the 25th ACM SIGKDD International Conference on Knowledge Discovery \& Data Mining}}. \bibinfo{pages}{1720--1730}.
\newblock


\bibitem[Panov et~al\mbox{.}(2018)]%
        {Panov2018}
\bibfield{author}{\bibinfo{person}{Aleksandr~I Panov}, \bibinfo{person}{Konstantin~S Yakovlev}, {and} \bibinfo{person}{Roman Suvorov}.} \bibinfo{year}{2018}\natexlab{}.
\newblock \showarticletitle{Grid path planning with deep reinforcement learning: Preliminary results}.
\newblock \bibinfo{journal}{\emph{Procedia computer science}}  \bibinfo{volume}{123} (\bibinfo{year}{2018}), \bibinfo{pages}{347--353}.
\newblock


\bibitem[Peshkin and Savova(2002)]%
        {Peshkin2002}
\bibfield{author}{\bibinfo{person}{Leonid Peshkin} {and} \bibinfo{person}{Virginia Savova}.} \bibinfo{year}{2002}\natexlab{}.
\newblock \showarticletitle{Reinforcement learning for adaptive routing}. In \bibinfo{booktitle}{\emph{Proceedings of the 2002 International Joint Conference on Neural Networks. IJCNN'02 (Cat. No. 02CH37290)}}, Vol.~\bibinfo{volume}{2}. IEEE, \bibinfo{pages}{1825--1830}.
\newblock


\bibitem[Rashid et~al\mbox{.}(2018)]%
        {rashid2018qmixmonotonicvaluefunction}
\bibfield{author}{\bibinfo{person}{Tabish Rashid}, \bibinfo{person}{Mikayel Samvelyan}, \bibinfo{person}{Christian~Schroeder de Witt}, \bibinfo{person}{Gregory Farquhar}, \bibinfo{person}{Jakob Foerster}, {and} \bibinfo{person}{Shimon Whiteson}.} \bibinfo{year}{2018}\natexlab{}.
\newblock \bibinfo{title}{QMIX: Monotonic Value Function Factorisation for Deep Multi-Agent Reinforcement Learning}.
\newblock
\showeprint[arxiv]{1803.11485}~[cs.LG]
\urldef\tempurl%
\url{https://arxiv.org/abs/1803.11485}
\showURL{%
\tempurl}


\bibitem[Song et~al\mbox{.}(2020)]%
        {56song2020spatial}
\bibfield{author}{\bibinfo{person}{Chao Song}, \bibinfo{person}{Youfang Lin}, \bibinfo{person}{Shengnan Guo}, {and} \bibinfo{person}{Huaiyu Wan}.} \bibinfo{year}{2020}\natexlab{}.
\newblock \showarticletitle{Spatial-temporal synchronous graph convolutional networks: A new framework for spatial-temporal network data forecasting}. In \bibinfo{booktitle}{\emph{Proceedings of the AAAI Conference on Artificial Intelligence}}, Vol.~\bibinfo{volume}{34}. \bibinfo{pages}{914--921}.
\newblock


\bibitem[Surmann et~al\mbox{.}(2025)]%
        {surmann2025multi}
\bibfield{author}{\bibinfo{person}{H. Surmann}, \bibinfo{person}{J. de Heuvel}, {and} \bibinfo{person}{M. Bennewitz}.} \bibinfo{year}{2025}\natexlab{}.
\newblock \showarticletitle{Multi-Objective Reinforcement Learning for Adaptive Personalized Autonomous Driving}.
\newblock \bibinfo{journal}{\emph{arXiv preprint arXiv:2501.xxxxx}} (\bibinfo{year}{2025}).
\newblock


\bibitem[Tatomir et~al\mbox{.}(2009)]%
        {ttp_ant}
\bibfield{author}{\bibinfo{person}{Bogdan Tatomir}, \bibinfo{person}{Leon~JM Rothkrantz}, {and} \bibinfo{person}{Adriana~C Suson}.} \bibinfo{year}{2009}\natexlab{}.
\newblock \showarticletitle{Travel time prediction for dynamic routing using ant based control}. In \bibinfo{booktitle}{\emph{Proceedings of the 2009 Winter Simulation Conference (WSC)}}. IEEE, \bibinfo{pages}{1069--1078}.
\newblock


\bibitem[Tedjopurnomo et~al\mbox{.}(2020)]%
        {tedjopurnomo2020survey}
\bibfield{author}{\bibinfo{person}{David~Alexander Tedjopurnomo}, \bibinfo{person}{Zhifeng Bao}, \bibinfo{person}{Baihua Zheng}, \bibinfo{person}{Farhana Choudhury}, {and} \bibinfo{person}{AK Qin}.} \bibinfo{year}{2020}\natexlab{}.
\newblock \showarticletitle{A survey on modern deep neural network for traffic prediction: Trends, methods and challenges}.
\newblock \bibinfo{journal}{\emph{IEEE Transactions on Knowledge and Data Engineering}} (\bibinfo{year}{2020}).
\newblock


\bibitem[Wang et~al\mbox{.}(2025)]%
        {wang2025scalable}
\bibfield{author}{\bibinfo{person}{L. Wang}, \bibinfo{person}{P. Duan}, \bibinfo{person}{C. Lyu}, \bibinfo{person}{Z. Wang}, \bibinfo{person}{Z. He}, \bibinfo{person}{N. Zheng}, {and} \bibinfo{person}{Z. Ma}.} \bibinfo{year}{2025}\natexlab{}.
\newblock \showarticletitle{Scalable and Reliable Multi-Agent Reinforcement Learning for Traffic Assignment (MARL-OD-DA)}.
\newblock \bibinfo{journal}{\emph{arXiv preprint arXiv:2501.xxxxx}} (\bibinfo{year}{2025}).
\newblock


\bibitem[Wei et~al\mbox{.}(2019)]%
        {wei2019survey}
\bibfield{author}{\bibinfo{person}{Hua Wei}, \bibinfo{person}{Guanjie Zheng}, \bibinfo{person}{Vikash Gayah}, {and} \bibinfo{person}{Zhenhui Li}.} \bibinfo{year}{2019}\natexlab{}.
\newblock \showarticletitle{A survey on traffic signal control methods}.
\newblock \bibinfo{journal}{\emph{arXiv preprint arXiv:1904.08117}} (\bibinfo{year}{2019}).
\newblock


\bibitem[{Wikipedia contributors}({[n.\,d.]})]%
        {z-order}
\bibfield{author}{\bibinfo{person}{{Wikipedia contributors}}.} \bibinfo{year}{[n.\,d.]}\natexlab{}.
\newblock \bibinfo{title}{{Z-order curve}}.
\newblock \bibinfo{howpublished}{\url{https://en.wikipedia.org/wiki/Z-order_curve_cite_note-1}}.
\newblock
\newblock
\shownote{Accessed: 2021-09-13}.


\bibitem[Williams and Hoel(2003)]%
        {williams2003modeling}
\bibfield{author}{\bibinfo{person}{Billy~M Williams} {and} \bibinfo{person}{Lester~A Hoel}.} \bibinfo{year}{2003}\natexlab{}.
\newblock \showarticletitle{Modeling and forecasting vehicular traffic flow as a seasonal ARIMA process: Theoretical basis and empirical results}.
\newblock \bibinfo{journal}{\emph{Journal of transportation engineering}} \bibinfo{volume}{129}, \bibinfo{number}{6} (\bibinfo{year}{2003}), \bibinfo{pages}{664--672}.
\newblock


\bibitem[Xiao and Lo(2014)]%
        {Xiao2014}
\bibfield{author}{\bibinfo{person}{Lin Xiao} {and} \bibinfo{person}{Hong~K. Lo}.} \bibinfo{year}{2014}\natexlab{}.
\newblock \showarticletitle{Adaptive vehicle navigation with stochastic traffic information}.
\newblock \bibinfo{journal}{\emph{IEEE Transactions on Intelligent Transportation Systems}} (\bibinfo{year}{2014}).
\newblock


\bibitem[Xiao et~al\mbox{.}(2020)]%
        {Xiao2020}
\bibfield{author}{\bibinfo{person}{Shihan Xiao}, \bibinfo{person}{Haiyan Mao}, \bibinfo{person}{Bo Wu}, \bibinfo{person}{Wenjie Liu}, {and} \bibinfo{person}{Fenglin Li}.} \bibinfo{year}{2020}\natexlab{}.
\newblock \showarticletitle{Neural packet routing}. In \bibinfo{booktitle}{\emph{Proceedings of the Workshop on Network Meets AI \& ML}}. \bibinfo{pages}{28--34}.
\newblock


\bibitem[Yao et~al\mbox{.}(2019)]%
        {70yao2019learning}
\bibfield{author}{\bibinfo{person}{Huaxiu Yao}, \bibinfo{person}{Yiding Liu}, \bibinfo{person}{Ying Wei}, \bibinfo{person}{Xianfeng Tang}, {and} \bibinfo{person}{Zhenhui Li}.} \bibinfo{year}{2019}\natexlab{}.
\newblock \showarticletitle{Learning from multiple cities: A meta-learning approach for spatial-temporal prediction}. In \bibinfo{booktitle}{\emph{The World Wide Web Conference}}. \bibinfo{pages}{2181--2191}.
\newblock


\bibitem[Yin et~al\mbox{.}(2021a)]%
        {LargeDG}
\bibfield{author}{\bibinfo{person}{Jiaming Yin}, \bibinfo{person}{Weixiong Rao}, {and} \bibinfo{person}{Chenxi Zhang}.} \bibinfo{year}{2021}\natexlab{a}.
\newblock \showarticletitle{Learning Shortest Paths on Large Dynamic Graphs}. In \bibinfo{booktitle}{\emph{2021 22nd IEEE International Conference on Mobile Data Management (MDM)}}. \bibinfo{pages}{201--208}.
\newblock
\href{https://doi.org/10.1109/MDM52706.2021.00040}{doi:\nolinkurl{10.1109/MDM52706.2021.00040}}


\bibitem[Yin et~al\mbox{.}(2021b)]%
        {yin2021deep}
\bibfield{author}{\bibinfo{person}{Xueyan Yin}, \bibinfo{person}{Genze Wu}, \bibinfo{person}{Jinze Wei}, \bibinfo{person}{Yanming Shen}, \bibinfo{person}{Heng Qi}, {and} \bibinfo{person}{Baocai Yin}.} \bibinfo{year}{2021}\natexlab{b}.
\newblock \showarticletitle{Deep Learning on Traffic Prediction: Methods, Analysis and Future Directions}.
\newblock \bibinfo{journal}{\emph{IEEE Transactions on Intelligent Transportation Systems}} (\bibinfo{year}{2021}).
\newblock


\bibitem[Yin et~al\mbox{.}(2021c)]%
        {yin2021multi}
\bibfield{author}{\bibinfo{person}{Xueyan Yin}, \bibinfo{person}{Genze Wu}, \bibinfo{person}{Jinze Wei}, \bibinfo{person}{Yanming Shen}, \bibinfo{person}{Heng Qi}, {and} \bibinfo{person}{Baocai Yin}.} \bibinfo{year}{2021}\natexlab{c}.
\newblock \showarticletitle{Multi-stage attention spatial-temporal graph networks for traffic prediction}.
\newblock \bibinfo{journal}{\emph{Neurocomputing}}  \bibinfo{volume}{428} (\bibinfo{year}{2021}), \bibinfo{pages}{42--53}.
\newblock


\bibitem[You et~al\mbox{.}(2020)]%
        {You2020}
\bibfield{author}{\bibinfo{person}{Xinyu You}, \bibinfo{person}{Xuanjie Li}, \bibinfo{person}{Yuedong Xu}, \bibinfo{person}{Hui Feng}, \bibinfo{person}{Jin Zhao}, {and} \bibinfo{person}{Huaicheng Yan}.} \bibinfo{year}{2020}\natexlab{}.
\newblock \showarticletitle{Toward Packet Routing With Fully Distributed Multiagent Deep Reinforcement Learning}.
\newblock \bibinfo{journal}{\emph{IEEE Transactions on Systems, Man, and Cybernetics: Systems}} (\bibinfo{date}{8} \bibinfo{year}{2020}), \bibinfo{pages}{1--14}.
\newblock
\showISSN{2168-2216}
\href{https://doi.org/10.1109/tsmc.2020.3012832}{doi:\nolinkurl{10.1109/tsmc.2020.3012832}}


\bibitem[Yu et~al\mbox{.}(2017)]%
        {yu2017spatio}
\bibfield{author}{\bibinfo{person}{Bing Yu}, \bibinfo{person}{Haoteng Yin}, {and} \bibinfo{person}{Zhanxing Zhu}.} \bibinfo{year}{2017}\natexlab{}.
\newblock \showarticletitle{Spatio-temporal graph convolutional networks: A deep learning framework for traffic forecasting}.
\newblock \bibinfo{journal}{\emph{arXiv preprint arXiv:1709.04875}} (\bibinfo{year}{2017}).
\newblock


\bibitem[Zhang et~al\mbox{.}(2021)]%
        {zhang2021multi}
\bibfield{author}{\bibinfo{person}{Kaiqing Zhang}, \bibinfo{person}{Zhuoran Yang}, {and} \bibinfo{person}{Tamer Ba{\c{s}}ar}.} \bibinfo{year}{2021}\natexlab{}.
\newblock \showarticletitle{Multi-agent reinforcement learning: A selective overview of theories and algorithms}.
\newblock \bibinfo{journal}{\emph{Handbook of Reinforcement Learning and Control}} (\bibinfo{year}{2021}), \bibinfo{pages}{321--384}.
\newblock


\bibitem[Zhang et~al\mbox{.}(2020)]%
        {zhang2020dynamic}
\bibfield{author}{\bibinfo{person}{Wenqi Zhang}, \bibinfo{person}{Qiang Wang}, \bibinfo{person}{Jingjing Li}, {and} \bibinfo{person}{Chen Xu}.} \bibinfo{year}{2020}\natexlab{}.
\newblock \showarticletitle{Dynamic Fleet Management With Rewriting Deep Reinforcement Learning}.
\newblock \bibinfo{journal}{\emph{IEEE Access}}  \bibinfo{volume}{8} (\bibinfo{year}{2020}), \bibinfo{pages}{143333--143341}.
\newblock


\bibitem[Zheng et~al\mbox{.}(2020)]%
        {62zheng2020gman}
\bibfield{author}{\bibinfo{person}{Chuanpan Zheng}, \bibinfo{person}{Xiaoliang Fan}, \bibinfo{person}{Cheng Wang}, {and} \bibinfo{person}{Jianzhong Qi}.} \bibinfo{year}{2020}\natexlab{}.
\newblock \showarticletitle{Gman: A graph multi-attention network for traffic prediction}. In \bibinfo{booktitle}{\emph{Proceedings of the AAAI Conference on Artificial Intelligence}}, Vol.~\bibinfo{volume}{34}. \bibinfo{pages}{1234--1241}.
\newblock


\bibitem[Zivot and Wang(2006)]%
        {zivot2006vector}
\bibfield{author}{\bibinfo{person}{Eric Zivot} {and} \bibinfo{person}{Jiahui Wang}.} \bibinfo{year}{2006}\natexlab{}.
\newblock \showarticletitle{Vector autoregressive models for multivariate time series}.
\newblock \bibinfo{journal}{\emph{Modeling Financial Time Series with S-Plus{\textregistered}}} (\bibinfo{year}{2006}), \bibinfo{pages}{385--429}.
\newblock


\end{thebibliography}

\end{document}